\documentclass[sigconf, nonacm]{acmart}
\AtBeginDocument{%
  }

\AtBeginDocument{%
  }

\usepackage[most]{tcolorbox}
\tcbuselibrary{listings,breakable,skins}
\usepackage{xcolor}

\definecolor{placeholderblue}{RGB}{0,90,180}
\definecolor{promptbg}{gray}{0.965}
\definecolor{promptframe}{gray}{0.78}

\newtcblisting{promptbox}{
  enhanced,
  breakable,
  listing only,
  colback=promptbg,
  colframe=promptframe,
  boxrule=0.35pt,
  arc=2mm,
  outer arc=2mm,
  left=1.8mm,
  right=1.8mm,
  top=1.5mm,
  bottom=1.5mm,
  before=\par\noindent,
  after=\par,
  listing options={
    basicstyle=\ttfamily\fontsize{9pt}{9.5pt}\selectfont,
    columns=fullflexible,
    keepspaces=false,
    showstringspaces=false,
    breaklines=true,
    breakatwhitespace=true,
    breakindent=0pt,
    lineskip=-2.0pt,
    breakautoindent=false,
    tabsize=2,
    escapeinside={(*@}{@*)},
    aboveskip=0pt,
    belowskip=0pt,
    xleftmargin=0pt,
    resetmargins=true
  }
}

\usepackage[most]{tcolorbox}
\tcbuselibrary{listings,breakable,skins}
\usepackage{xcolor}

\definecolor{promptbg}{gray}{0.965}
\definecolor{promptframe}{gray}{0.25}
\definecolor{prompttitlebg}{gray}{0.18}

\newtcblisting{prompttemplate}[1]{
  enhanced,
  breakable,
  listing only,
  colback=promptbg,
  colframe=prompttitlebg,
  boxrule=1.1pt,
  colbacktitle=prompttitlebg,
  coltitle=white,
  title={#1},
  fonttitle=\bfseries\small,
  arc=1.5mm,
  outer arc=1.5mm,
  left=1.8mm,
  right=1.8mm,
  top=1.5mm,
  bottom=1.5mm,
  before skip=6pt,
  after skip=8pt,
  listing options={
    basicstyle=\ttfamily\fontsize{8.8pt}{9.5pt}\selectfont,
    columns=fullflexible,
    keepspaces=false,
    showstringspaces=false,
    breaklines=true,
    breakatwhitespace=true,
    breakindent=0pt,
    breakautoindent=false,
    escapeinside={(*@}{@*)},
    tabsize=2,
    aboveskip=0pt,
    belowskip=0pt,
    xleftmargin=0pt,
    resetmargins=true
  }
}

\usepackage{multirow}
\usepackage{graphicx}
\usepackage{booktabs}
\usepackage{enumitem}
\usepackage{subcaption}
\usepackage{makecell}
\usepackage{algorithm}
\usepackage{algpseudocode}
\usepackage{placeins}
\newcommand{\std}[2]{#1\lower0.05ex\hbox{\fontsize{6.5}{0}\selectfont$\pm$#2}}
\newcommand{\stdb}[2]{\textbf{#1}\lower0.05ex\hbox{\fontsize{6.5}{0}\selectfont\textbf{$\pm$#2}}}
\newcommand{\stdu}[2]{\underline{#1\lower0.05ex\hbox{\fontsize{6.5}{0}\selectfont$\pm$#2}}}

\begin{document}


\title{L2IR: Revealing Latent Intent in Graph Fraud Detection}
\author{Jinsheng Guo}
\authornote{Equal Contribution}
\email{guojinsheng@mail.hfut.edu.cn}
\affiliation{%
  \institution{Hefei University of Technology}
  \city{Hefei}
  \country{China}
}

\author{Zhenhao Weng}
\authornotemark[1]
\email{zhenhaoweng@mail.hfut.edu.cn}
\affiliation{%
  \institution{Hefei University of Technology}
  \city{Hefei}
  \country{China}
}

\author{Yibo Liu}
\email{yibo.liu@mail.hfut.edu.cn}
\affiliation{%
  \institution{Hefei University of Technology}
  \city{Hefei}
  \country{China}
}

\author{Yan Qiao}
\authornote{Corresponding authors}
\email{qiaoyan@hfut.edu.cn}
\affiliation{%
  \institution{Hefei University of Technology}
  \city{Hefei}
  \country{China}
}

\author{Meng Li}
\authornotemark[2]
\email{mengli@hfut.edu.cn}
\affiliation{%
  \institution{Hefei University of Technology}
  \city{Hefei}
  \country{China}
}

\begin{abstract}
Graph fraud detection has long depended on Graph Neural Networks (GNNs) to propagate and aggregate information across relational data. A critical obstacle in practice, however, is that fraudsters frequently disguise themselves by forging numerous connections with benign users, causing fraud signals to be progressively diluted during neighborhood aggregation and undermining detection reliability. While recent efforts have used Large Language Models (LLMs) to provide rich semantic cues for fraud detection, the underlying intent behind suspicious connections remains insufficiently explored. Compounding this issue, the scarcity of annotated fraud samples makes it difficult to train detectors that remain robust under heavy camouflage. To address these gaps, we propose L2IR, an \textbf{L}LM-driven \textbf{L}atent \textbf{I}ntent \textbf{R}evealing framework for graph fraud detection. By uncovering latent intent from both user behaviors and suspicious connections, L2IR extracts intent-aware representations from raw behavioral traces and reasons about the true purpose behind individual connections, effectively distinguishing supportive links from misleading ones. It further incorporates adaptive self-training to enhance robustness under limited supervision. Evaluations on two real-world datasets characterized by pervasive camouflage demonstrate that L2IR surpasses strong baselines and can function as a plug-in enhancement for a range of GNN-based detectors, improving AUPRC by up to 8.27\%.
\end{abstract}



\maketitle
\section{Introduction}
Modern online information systems, such as e-commerce platforms, review communities, social networks, and financial services, generate rich relational data among users, items, content, and transactions \cite{akoglu2015graph,rayana2015collective,weng2019cats,wang2019semi}. Fraudulent activities in these systems reduce service quality and weaken system trust, making fraud detection an important task \cite{rayana2015collective,lu2022bright}. In response, graph-based methods have become an effective paradigm. Graph-based methods model entities and interactions as graphs and exploit both node attributes and relational structure for detection \cite{akoglu2015graph,hooi2016fraudar}. Among them, Graph Neural Networks (GNNs) are widely used, since they learn center node representations by aggregating information from connected neighbors, thereby capturing patterns from neighboring nodes that are critical for distinguishing fraudulent entities from benign ones \cite{kipf2016semi,hamilton2017inductive,velivckovic2018graph}. 

\begin{figure}[t]
    \centering
    \includegraphics[width=\linewidth]{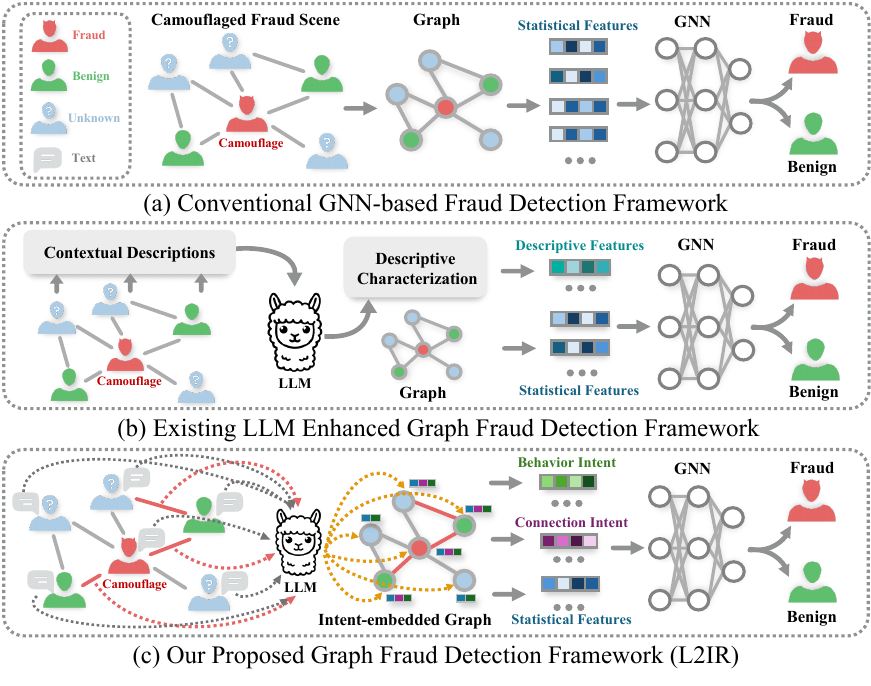}
    \caption{Comparison of our proposed L2IR with GNN-based and LLM enhanced GNN methods under camouflage. (a) Standard GNN-based methods rely solely on graph structure. (b) LLM enhanced GNN methods incorporate dataset-level semantic descriptions to improve performance. (c) Our L2IR analyzes the intent behind each user interaction, enabling effective identification of camouflaged fraud. }
    \label{fig:motivation}
\end{figure}

However, existing GNN-based fraud detection methods mainly rely on neighborhood aggregation over observed graph structures (as illustrated in Fig.~\ref{fig:motivation}(a) \cite{huang2025can,yang2025flag}) and insufficiently explore fraudsters' camouflage behaviors \cite{hooi2016fraudar,dou2020enhancing}. In relation camouflage scenarios \cite{dou2020enhancing}, fraudsters build many connections with benign users while conducting fraudulent activities on only a small fraction of them, so the benign-dominated neighborhood dilutes fraud signals during aggregation, leading to less reliable detection and increased false negatives \cite{dou2020enhancing,liu2020alleviating}.

Recently, driven by the context reasoning and semantic comprehension capabilities of Large Language Models (LLMs) \cite{he2024harnessing,tang2024graphgpt,zhu2024touchup}, existing studies leverage LLMs to incorporate rich semantics to assist graph fraud detection \cite{huang2025can,yang2025flag,li2026dgp}. These studies have revealed potential opportunities for detecting camouflaged fraud by integrating descriptive semantic characterization with graph structural information (as illustrated in Fig.~\ref{fig:motivation}(b)). However, existing methods primarily restrict LLMs to generating global semantic descriptions of the dataset, without reasoning about the intent behind connections. Consequently, they fail to fully exploit the potential of LLMs for recognizing camouflaged fraud, still suffering from notable misclassifications. Even worse, real-world camouflaged fraud data is often constrained by limited supervision: the strong stealthiness of camouflaged fraud makes it extremely difficult to label such cases in datasets, leaving insufficient supervision for robust training and generalization \cite{wang2019semi,liu2021pick,yu2024barely}.

To tackle these challenges, we propose \textbf{L2IR}, an \textbf{L}LM-driven \textbf{L}atent \textbf{I}ntent \textbf{R}evealing framework for graph fraud detection. Specifically, L2IR leverages LLMs to analyze the behavioral semantics of each user to infer the intent behind individual connections, and then incorporates this inferred intent information into node features, thereby helping to uncover deeply camouflaged fraudulent nodes (as illustrated in Fig.~\ref{fig:motivation}(c)). In addition, L2IR adopts an adaptive self-training mechanism to augment supervision with reliable signals from previous stages, thereby improving robustness under limited fraud labels. Extensive experiments on two real-world datasets demonstrate the effectiveness of L2IR in fraud detection, with clear gains under camouflage and limited supervision scenarios.

In summary, our main contributions are as follows:

\begin{itemize}
\item We introduce a novel LLM-driven perspective for graph fraud detection under camouflage by inferring the intent behind suspicious connections with LLMs, which effectively improves the reliability of graph fraud detection.
\item We propose L2IR, an intent modeling framework for graph fraud detection, which jointly models behavior intent and connection intent to extract deep semantic features. It also integrates adaptive self-training to address the scarcity of fraud labels, which enables accurate detection of heavily camouflaged fraud even under limited supervision.
\item We evaluate L2IR against nine representative baselines on two real-world datasets. The results demonstrate that L2IR achieves superior empirical performance and can also serve as a plug-in to improve existing GNN-based fraud detectors.
\end{itemize}

\section{Related Works}\label{sec:related-work}

\subsection{Graph Neural Networks for Fraud Detection}
GNNs provide a powerful framework for capturing fraudulent patterns by aggregating neighborhood information through message propagation mechanisms \citep{liu2018heterogeneous, liu2019geniepath, shi2022h2,wang2020gcn}. Early works directly applied classic GNNs like Graph Convolutional Networks (GCN) \citep{kipf2016semi} and Graph Attention Networks (GAT) \citep{velivckovic2018graph} to fraud detection. Subsequent works introduced various enhancements.
GraphSAGE \citep{hamilton2017inductive} introduces an inductive framework that learns to generate node embeddings by sampling and aggregating features from a node's local neighborhood, allowing effective embedding creation for unseen nodes in dynamic or new graphs.
RGTAN \citep{xiang2025enhancing} models transaction sequences as temporal graphs and uses gated attention to capture temporal fraud patterns. It also learns neighbor risk representations to detect multi-hop fraud structures.
DiffGraph \citep{li2025diffgraph} employs a latent diffusion paradigm to filter noise and captures relation transitions through a bidirectional diffusion process for better node representations. To combat label scarcity, some unsupervised methods employ mutual information maximization or contrastive learning \citep{velivckovic2018deep, zhu2020deep}, while SemiGNN \citep{wang2019semi} links labeled and unlabeled users via social relations and integrates heterogeneous data sources with hierarchical attention for fraud detection.
However, these methods do not exploit the rich textual information and struggle to cope with the challenges posed by fraudsters deliberately camouflaging themselves as normal nodes \citep{hooi2016fraudar, liu2021pick, wang2019fdgars}.

\subsection{Integrating LLMs with GNNs for Fraud Detection}
In recent years, LLMs have demonstrated powerful language understanding \citep{minaee2024large, thirunavukarasu2023large,chang2024survey,achiam2023gpt}, opening a new path for graph learning by exploiting the rich textual semantics of nodes \citep{wei2024llmrec, tan2023walklm,zhao2022learning}. For instance, one paradigm utilizes GNNs to enhance LLMs, where LLMs serve as the primary predictors. GraphGPT \citep{tang2024graphgpt} employs graph instruction tuning to translate graph structures into LLM-compatible representations, while InstructGLM \citep{ye2024language} relies on natural language prompts for the same purpose. In both cases, the goal is to enable direct graph learning with LLMs, eliminating the need for task-specific GNN fine-tuning. DGP \citep{li2026dgp} proposes a dual granularity prompting framework retaining fine-grained text for target nodes while summarizing neighbors into coarse-grained prompts, reducing input size while preserving key fraud semantics. However, employing LLMs as the final predictor imposes significant computational resource demands and suffers from low inference efficiency. Conversely, an alternative paradigm employs LLMs to enhance GNNs, with GNNs serving as the primary predictors. TAPE \citep{he2024harnessing} pioneers this by using an LLM to generate explanations as augmented node features. TouchUp-G \citep{zhu2024touchup} further enhances node features from pre-trained models through graph-centric finetuning to better align them with the graph structure. MLED \citep{huang2025can} uses type-level and relation-level enhancers to integrate text knowledge with graph structure for better fraud-benign distinction. FLAG \citep{yang2025flag} employs semantic similarity sampling to filter camouflaged neighbors and leverages LLMs to extract discriminative textual features, thereby improving fraud detection, particularly on text-rich graphs. Although LLM enhanced GNNs mitigate the high cost of the LLM-as-Predictor paradigm, they fail to explicitly capture the true intent behind camouflaged connections, leaving misleading links unrecognized and contributing to elevated false negatives. Moreover, these methods still struggle under label scarcity due to limited supervision.

In contrast to prior work, L2IR leverages LLMs to infer the intent behind behaviors and connections, enhancing GNNs' capacity to detect camouflaged fraud.

\begin{figure*}[t]
  \centering
  \includegraphics[width=\textwidth]{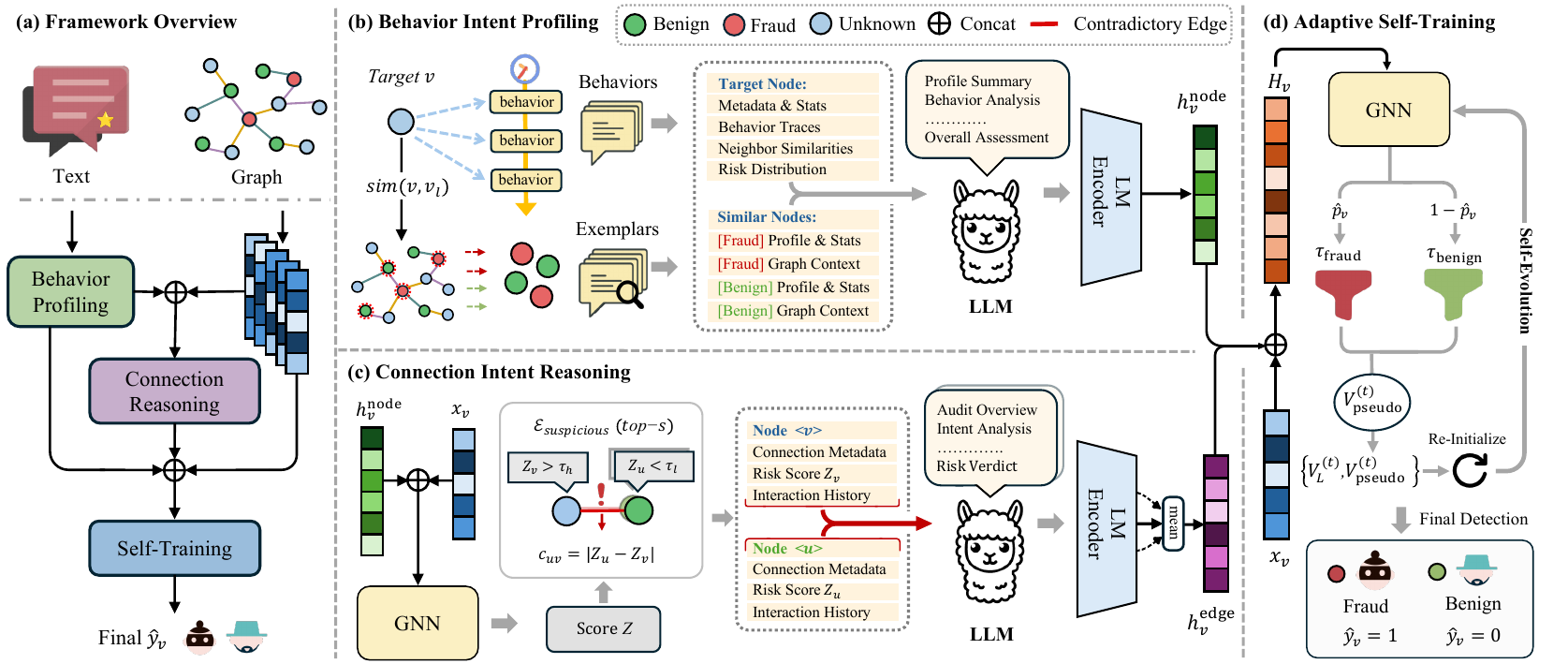}
  \caption{The overall framework of L2IR. It consists of three key modules: (b) Behavior Intent Profiling generates intent profiles; (c) Connection Intent Reasoning infers the intent behind each connection; (d) Adaptive Self-Training progressively augments training labels to handle label scarcity.}
  \label{fig:framework}
\end{figure*}

\section{Preliminaries}

We consider graph-based fraud detection on a heterogeneous graph $\mathcal{G} = (\mathcal{V}, \mathcal{E}, \mathcal{R})$, where $\mathcal{V}$ is the set of $N$ nodes, $\mathcal{E} \subseteq \mathcal{V} \times \mathcal{V} \times \mathcal{R}$ is the set of typed edges, and $\mathcal{R}$ is the set of relation types. Each node $v$ is associated with a statistical feature vector $x_v \in \mathbb{R}^{d}$ derived from metadata, and a raw textual sequence $T_v$ representing its historical behavior traces (e.g., textual reviews, item ratings, and interaction timestamps). These records are key to uncovering semantic intent hidden behind camouflaged connections. The neighborhood of node $v$ under relation $r$ is denoted as $\mathcal{N}_r(v) = \{u \mid (u, v, r) \in \mathcal{E}\}$. We further define the homogeneous projection $\mathcal{G}_{\text{homo}} = (\mathcal{V}, \mathcal{E}_{\text{homo}})$, where $\mathcal{E}_{\text{homo}} = \bigcup_{r \in \mathcal{R}} \{(u,v) \mid (u,v,r) \in \mathcal{E}\}$ collapses all specific relations (e.g., reviewing the same products or giving identical ratings) into an untyped edge set to represent basic structural connectivity.

We formulate our target scenario as a graph fraud detection task operating under annotation scarcity. Formally, let $\mathcal{V}_L \subset \mathcal{V}$ be a small set of labeled nodes with ground-truth labels $y_v \in \{0, 1\}$ (where $1$ denotes a fraudster and $0$ a benign entity), and $\mathcal{V}_U = \mathcal{V} \setminus \mathcal{V}_L$ be the remaining unlabeled nodes. In this setting, the model observes the complete topology $\mathcal{G}$ and all node statistical features $\mathcal{X} = \{x_v\}_{v \in \mathcal{V}}$ during training, but is supervised only by $\mathcal{V}_L$. Our objective is to learn a predictive function $f_\theta: \mathcal{V} \to \{0, 1\}$ that accurately uncovers all hidden fraudsters without triggering false positives on incidentally connected benign users.

\section{Methodology}
In this section, we present the technical details of the proposed L2IR framework. It integrates LLM-driven semantic reasoning with graph structural information by modeling both node behavior intent and edge connection intent. We first outline the overall architecture, and then provide a detailed description of the constituent modules.

\subsection{Framework Overview}
\label{sec:overview}

At a high level, L2IR uncovers camouflaged fraudsters by aligning graph structure with intent features. As illustrated in Figure~\ref{fig:framework}, L2IR consists of three key modules: \textit{\textbf{1) Behavior Intent Profiling:}} it prompts an LLM to digest raw chronological behavior traces and retrieved exemplars, and yields node-level behavior intent representations ($h_v^{\text{node}}$) that complement statistical features with behavior semantic evidence. \textit{\textbf{2) Connection Intent Reasoning:}} this module applies a preliminary GNN to score nodes and flag suspicious connections, then prompts an LLM to cross-audit each one, reasoning about the true intent behind the interaction. This generates edge-level connection intent representations ($h_v^{\text{edge}}$) that help distinguish supportive connections from misleading ones. \textit{\textbf{3) Adaptive Self-Training:}} under severe supervision scarcity, this module deploys an iterative self-training loop tailored to fraud detection. Using asymmetric confidence thresholds, it progressively adds high-confidence pseudo-labels to the training set, preventing early-round errors from cascading into later ones. Through this pipeline, semantic reasoning and structural learning reinforce each other to robustly identify camouflaged fraudsters.

\subsection{Behavior Intent Profiling}
\label{sec:Behavior-profiling}
To infer behavior intent from raw traces $T_v$, we prompt an LLM to analyze each target node against a group of retrieved exemplars of labeled fraud and benign nodes, allowing the model to recognize latent intent by identifying which behavioral patterns align with these exemplars.

\textbf{Dynamic Exemplar Retrieval.} To provide the LLM with references for behavior intent profiling, we dynamically retrieve a few exemplars from the labeled nodes for each target node \(v\). Specifically, we compute the similarity between \(v\) and each labeled node \(v_l \in \mathcal{V}_L\) using a combined metric:
\begin{equation}
  \operatorname{sim}(v, v_l) =
  \alpha \cdot \operatorname{sim}_{\text{text}}(q_v, q_{v_l})
  + (1 - \alpha) \cdot \operatorname{sim}_{\text{inter}}(I_v, I_{v_l}),
\end{equation}
where \(q_v\) denotes the textual vector constructed from \(T_v\), \(I_v\) denotes the historically interacted item set of node \(v\), \(\operatorname{sim}_{\text{text}}\) is the cosine similarity between textual vectors, and \(\operatorname{sim}_{\text{inter}}\) is the Jaccard index between item sets, where a higher value indicates a greater degree of interaction. The coefficient \(\alpha \in [0, 1]\) balances semantic content and behavioral patterns.

Based on the similarities, we construct an exemplar pool for target node $v$. For each class $c \in \{0,1\}$, we rank nodes $v_l \in \mathcal{V}_L^c \setminus \{v\}$ according to the similarity score $\operatorname{sim}(v, v_l)$, and retrieve the top-$k$ most similar nodes to form the class-specific exemplar set $\mathcal{S}_v^c$. The integrated exemplar set $\mathcal{S}_v = \mathcal{S}_v^0 \cup \mathcal{S}_v^1$ serves as a contrastive basis, which provides behaviorally grounded references for the LLM to effectively distinguish benign from fraudulent behavior intent.

\textbf{Behavior Intent Profiling with LLM.} We design a prompt to guide a pre-trained LLM, denoted by $\mathcal{M}_{\text{LLM}}$, in reasoning over the target node's behavior traces $T_v$ and the labeled exemplars $\mathcal{S}_v$, yielding a structured profile $b_v = \mathcal{M}_{\text{LLM}}(\texttt{Prompt}_{\text{profile}}(v))$. 

As shown in Table~\ref{tab:prompt_example}, the prompt consists of four components: \textit{Role}, \textit{Exemplars}, \textit{Target Node} and \textit{Output}. The \textit{Role} anchors the LLM in the fraud detection domain, reducing irrelevant reasoning. The \textit{Exemplars} provides $k$ pairs of fraud-benign cases with consistent metadata, offering contrastive behavioral references that guide the LLM toward patterns relevant to intent inference. The \textit{Target Node} supplies the node's statistical data, graph relation context, and chronological behavior traces, providing the LLM with concrete behavioral evidence to analyze. The \textit{Output} enforces a structured schema progressing from profile summary to overall assessment, making the resulting profile $b_v$ suitable for downstream encoding. The complete template for behavior intent profiling is provided in Appendix~\ref{app:complete_prompt}.

\begin{table}[t]
  \centering
  \caption{Core format of prompt for Behavior Intent Profiling.}
  \label{tab:prompt_example}
  \small
  \renewcommand{\arraystretch}{1.2}
  \begin{tabular}{@{}p{0.95\columnwidth}@{}}
    \toprule
    \textbf{[Role]} Domain-expert persona: senior fraud detection analyst specializing in review behavior and relation camouflage. The core task is to infer the target node's behavior intent by strictly following all provided constraints and analytical requirements. \\
    \midrule
    \textbf{[Exemplars]} \\
    \textit{$<$$k$ fraudulent node cases: node metadata, review statistics, rating distribution, sentiment score, graph relation context, and ground-truth label.$>$} \\
    \textit{$<$$k$ benign node cases: same format as above.$>$} \\
    \midrule
    \textbf{[Target Node]} \\
    Node ID: \texttt{$<$uid$>$} \quad Total reviews: \texttt{$<$n$>$} \quad Avg. rating: \texttt{$<$r$>$} \\
    Graph Relation Context: \texttt{[Neighbor Metadata $|$ Behavior Similarities $|$ Risk Distribution]} \\
    Chronological Behavior Traces: \texttt{[Product $|$ Star Rating $|$ Text Content $|$ Helpfulness Score]} \\
    \midrule
    \textbf{[Output]} \textit{1) Node Profile Summary} $\to$ \textit{2) Behavior Pattern Analysis} $\to$ \textit{3) Fraud Signal Analysis} $\to$ \textit{4) Overall Assessment} \\
    \bottomrule
  \end{tabular}
\end{table}

\textbf{Semantic Encoding.}
To integrate the captured behavior intent into node features, we encode the structured output $b_v$ into a dense vector using a frozen pre-trained language encoder $\mathcal{M}_{\text{Enc}}$:
\begin{equation}
  h_v^{\text{node}} = \mathcal{M}_{\text{Enc}}(b_v) \in \mathbb{R}^{d_s}.
\end{equation}
The resulting embedding $h_v^{\text{node}}$ complements the statistical features $x_v$ as a node-level semantic prior, integrating behavior intent before neighborhood aggregation and equipping the downstream module with richer evidence for connection intent reasoning.

\subsection{Connection Intent Reasoning}
\label{sec:intent-reasoning}

To reason the connection intent in the graph structure, we first apply a preliminary GNN to score nodes and flag suspicious connections, then prompt an LLM to cross-audit each one by reasoning over both endpoints' interaction histories, allowing the model to identify intent signals that distinguish supportive connections.

\textbf{Suspicious Connection Detection.}
We train a preliminary GNN on the enhanced features $[x_v \| h_v^{\text{node}}]$ to produce fraud risk scores for all nodes. To enhance the robustness of scoring, we split the training set into $K$ distinct folds and train a separate model on each fold. For labeled nodes $\mathcal{V}_L$, we adopt an out-of-fold (OOF) strategy. Specifically, each node falls into exactly one validation fold and is scored exclusively by that fold's model; this prevents the bias from the remaining $K-1$ models that observed its label during training, thereby yielding an unbiased prediction $\hat{p}_v^{\text{oof}}$. For unlabeled nodes $\mathcal{V}_U$, since no ground-truth label is available, all $K$ fold models provide unbiased predictions; we therefore denote by $\hat{p}_v^{(k)}$ the prediction from the $k$-th fold model, and average these $K$ predictions to improve stability. The final risk score $Z_v$ is defined as:
\begin{equation}
Z_v = \begin{cases}
    \hat{p}_v^{\text{oof}}, & v \in \mathcal{V}_L \\[4pt]
    \dfrac{1}{K}\displaystyle\sum_{k=1}^{K}\hat{p}_v^{(k)}, & v \in \mathcal{V}_U
\end{cases}
\end{equation}

\begin{table}[b]
  \centering
  \caption{Core format of prompt for Connection Intent Reasoning.}
  \label{tab:audit_prompt_example}
  \small
  \renewcommand{\arraystretch}{1.2}
  \begin{tabular}{@{}p{0.95\columnwidth}@{}}
    \toprule
    \textbf{[Role]} Senior fraud audit analyst specializing in review graphs and relation camouflage. Node $v$/$u$ roles
    (\textit{Suspected Fraud Node} or \textit{Suspected Benign Node}) are dynamically assigned
    based on their relative preliminary risk scores $Z$, with the
    higher-scored node explicitly designated as the suspected side to analyze intent and assess evidence. \\
    \midrule
    \textbf{[Target Connection]} \\
    Connection Metadata: \texttt{[Node IDs \& Roles $|$ Risk scores: $\langle Z_v \rangle$ / $\langle Z_u \rangle$ $|$ Contradictory Magnitude: $\langle c_{uv} \rangle$]} \\
    Chronological Interaction History ($v$ \& $u$): \texttt{[Product $|$ Date $|$ Star Rating $|$ Text Content $|$ Helpfulness Score]} \\
    \midrule
    \textbf{[Output]} \textit{1) Connection Overview}
    (risk scores, suspicious, fraud signals) $\to$
    \textit{2) Behavior Difference} ($v$/$u$ rating and review divergence) $\to$
    \textit{3) Connection Intent Analysis} $\to$
    \textit{4) Counter Evidence and Uncertainty} $\to$
    \textit{5) Risk Verdict} (Low/Med/High $+$ confidence $+$ key evidence) \\
    \bottomrule
  \end{tabular}
\end{table}

Given the node risk scores $Z_v$, we partition the nodes into a suspected fraud set $\mathcal{V}_{+}$ and a suspected benign set $\mathcal{V}_{-}$ using dual thresholds $\tau_h$ and $\tau_l$ ($\tau_h > \tau_l$):
\begin{equation}
  \begin{aligned}
    \mathcal{V}_{+} &= \{v \in \mathcal{V} \mid Z_v > \tau_h\}, \\
    \mathcal{V}_{-} &= \{v \in \mathcal{V} \mid Z_v < \tau_l\}.
  \end{aligned}
\end{equation}
Based on this partition, we identify all edges from $\mathcal{E}_{\text{homo}}$ whose two endpoints belong to different node sets---one in $\mathcal{V}_{+}$ and the other in $\mathcal{V}_{-}$---and define them as the \textit{contradictory edge set}, denoted by $\mathcal{E}_{\text{contra}}$:
\begin{equation}
  \mathcal{E}_{\text{contra}} = \mathcal{E}_{\text{homo}} \cap \Big( (\mathcal{V}_{+} \times \mathcal{V}_{-}) \cup (\mathcal{V}_{-} \times \mathcal{V}_{+}) \Big).
\end{equation}
Next, we compute the \textit{contradictory magnitude} $c_{uv} = |Z_u - Z_v|$ for each contradictory edge and retain the top-$s$ edges to form $\mathcal{E}_{\text{suspicious}}$. The budget $s$ strictly bounds the LLM computational cost, focusing our reasoning directly on the deceptive boundaries that trigger false positives/negatives during GNN message passing.

\textbf{LLM Cross-Audit.}
For each suspicious edge $(u, v) \in \mathcal{E}_{\text{suspicious}}$, we design a prompt incorporating the interaction histories of both endpoints alongside their risk scores $Z_u$, $Z_v$ and \textit{contradictory magnitude} $c_{uv}$. As shown in Table~\ref{tab:audit_prompt_example}, the prompt consists of three components: \textit{Role}, \textit{Target Connection} and \textit{Output}. The \textit{Role} is dynamically assigned based on relative risk scores, so that the LLM can focus its reasoning on the right target. The \textit{Target Connection} provides both the quantitative risk context and the chronological interaction histories, grounding the audit in structural evidence. The \textit{Output} progresses from connection overview to a final risk verdict, ensuring the reasoning is systematic and the resulting report $r_{uv}$ captures the assessed connection intent. The complete template is provided in Appendix~\ref{app:complete_prompt}.

\textbf{Connection Intent Encoding.}
We encode the audit report $r_{uv}$ into a continuous representation
using the language encoder $\mathcal{M}_{\text{Enc}}$:
\begin{equation}
  e_{uv} = \mathcal{M}_{\text{Enc}}(r_{uv}) \in \mathbb{R}^{d_s}.
\end{equation}
This maps the LLM's edge-level reasoning into the GNN's embedding space, enabling downstream aggregation to suppress deceptive message propagation.

\textbf{Feature Fusion.} 
To incorporate the edge-level intent embeddings $e_{uv}$ into node-level features, we perform mean pooling over all suspicious edges connected to $v$:
\begin{equation}
  h_v^{\text{edge}} = \begin{cases}
  \frac{1}{|\mathcal{N}_{\text{suspicious}}(v)|} \sum_{u \in \mathcal{N}_{\text{suspicious}}(v)} e_{uv}, & \text{if } \mathcal{N}_{\text{suspicious}}(v) \neq \emptyset \\
  \mathbf{0}, & \text{otherwise}
  \end{cases}
\end{equation}
where $\mathcal{N}_{\text{suspicious}}(v) = \{u \mid (u, v) \in \mathcal{E}_{\text{suspicious}}\}$. This mean pooling distills the relational intent from suspicious connections, while for nodes without suspicious edges, we set $h_v^{\text{edge}} = \mathbf{0}$ to avoid introducing noise. Finally, we fuse the multi-view signals via concatenation:
\begin{equation}
  H_v = [x_v \| h_v^{\text{node}} \| h_v^{\text{edge}}] \in \mathbb{R}^{d + 2d_s}.
\end{equation}
The three components $x_v$, $h_v^{\text{node}}$, and $h_v^{\text{edge}}$ together cover statistical features, node-level behavior intent, and edge-level connection intent, forming a comprehensive node representation.

\subsection{Adaptive Self-Training}
\label{sec:self-training}
To address supervision scarcity under class imbalance, we deploy a self-training loop that progressively expands the training set with high-confidence pseudo-labels. At each round $t$, we first re-initialize the GNN, then train it on the current labeled set $\mathcal{V}_L^{(t)}$ with the enriched features $H_v$, and use it to generate pseudo-labels for the next round.

\textbf{Re-Initialization of GNN.} At the start of the $t$-th training round, the GNN is re-initialized from scratch. Early models trained on limited labels tend to misclassify camouflaged fraudsters as benign due to structural camouflage. If the model parameters $\theta$ are carried over iteratively, these biases can propagate through multi-hop message passing and compound across rounds. Re-initializing at each round forces the model to relearn from the updated labeled set, allowing it to progressively correct earlier misclassifications as more reliable pseudo-labels are incorporated.

\textbf{GNN Training.}
With freshly initialized parameters, the GNN is trained on $\mathcal{G}_{\text{homo}}$ using the enriched features $H_v$ and the current labeled set $\mathcal{V}_L^{(t)}$. The model minimizes the binary cross-entropy loss:
\begin{equation}
  \mathcal{L} = -\frac{1}{|\mathcal{V}_L^{(t)}|} \sum_{v \in
  \mathcal{V}_L^{(t)}} \left[ y_v \log \hat{p}_v + (1 - y_v)
  \log(1 - \hat{p}_v) \right],
\end{equation}
where $\hat{p}_v$ is the fraud probability for node $v$ predicted by the current model. Once training converges, the model produces $\hat{p}_v$ for all nodes in $\mathcal{V}_U$ for pseudo-label generation.

\begin{algorithm}[t]
\caption{The overall procedure of L2IR framework.}
\label{alg:L2IR}
\begin{algorithmic}[1]
\Require Heterogeneous graph $\mathcal{G} = (\mathcal{V}, \mathcal{E}, \mathcal{R})$, statistical features $\mathcal{X}$, textual traces $\{T_v\}_{v \in \mathcal{V}}$, labeled set $\mathcal{V}_L$, risk thresholds $\tau_h, \tau_l$, confidence thresholds $\tau_{\text{fraud}}, \tau_{\text{benign}}$, maximum self-training rounds $T$
\Ensure Trained model $f_\theta$

\Statex \textbf{Behavior Intent Profiling:}
\For{each node $v \in \mathcal{V}$}
    \State Retrieve exemplars $\mathcal{S}_v$ via textual and interaction similarity
    \State Prompt $\mathcal{M}_{\text{LLM}}$ with $T_v$ and $\mathcal{S}_v$ to generate profile $b_v$
    \State Encode $b_v$ to semantic embedding $h_v^{\text{node}} \gets \mathcal{M}_{\text{Enc}}(b_v)$
\EndFor

\Statex \textbf{Connection Intent Reasoning:}
\State Train preliminary GNN on $[x_v \| h_v^{\text{node}}]$ to get risk scores $Z_v$
\State Partition nodes into $\mathcal{V}_{+}$ and $\mathcal{V}_{-}$ via thresholds $\tau_h, \tau_l$
\State Retain top-$s$ contradictory edges ranked by score gap as $\mathcal{E}_{\text{suspicious}}$
\For{each $(u, v) \in \mathcal{E}_{\text{suspicious}}$}
    \State Cross-audit $(u,v)$ via $\mathcal{M}_{\text{LLM}}$ to generate audit report $r_{uv}$
    \State Encode audit to relational embedding $e_{uv} \gets \mathcal{M}_{\text{Enc}}(r_{uv})$
\EndFor
\State Mean-pool $\{e_{uv}\}$ to obtain edge representation $h_v^{\text{edge}}$
\State Concatenate the features: $H_v \gets [x_v \| h_v^{\text{node}} \| h_v^{\text{edge}}]$

\Statex \textbf{Adaptive Self-Training:}
\State Initialize labeled set $\mathcal{V}_L^{(0)} \gets \mathcal{V}_L$
\For{$t = 0, 1, \dots, T-1$}
    \State Re-initialize parameters $\theta$ from scratch
    \State Update $f_\theta$ by training on $H_v$ with $\mathcal{V}_L^{(t)}$
    \State Generate pseudo-labels $\hat{\mathcal{V}}_1, \hat{\mathcal{V}}_0$ via thresholds $\tau_{\text{fraud}}, \tau_{\text{benign}}$
    \State $\mathcal{V}_L^{(t+1)} \gets \mathcal{V}_L^{(t)} \cup \hat{\mathcal{V}}_1 \cup \hat{\mathcal{V}}_0$
\EndFor
\State \Return Trained model $f_\theta$
\end{algorithmic}
\end{algorithm}

\textbf{Pseudo-Label Generation.} 
Since $\hat{p}_v$ represents the model's confidence that node $v$ is fraudulent, $1 - \hat{p}_v$ naturally serves as the confidence for the benign class. Accordingly, we identify a pseudo-fraud set $\hat{\mathcal{V}}_1$ and a pseudo-benign set $\hat{\mathcal{V}}_0$ from the unlabeled nodes $\mathcal{V}_U$ by applying asymmetric confidence thresholds ($\tau_{\text{benign}} > \tau_{\text{fraud}}$):
\begin{equation}
  \begin{aligned}
    \hat{\mathcal{V}}_1 &= \{v \in \mathcal{V}_U \mid \hat{p}_v \geq \tau_{\text{fraud}}\}, \\
    \hat{\mathcal{V}}_0 &= \{v \in \mathcal{V}_U \mid 1 - \hat{p}_v \geq \tau_{\text{benign}}\}.
  \end{aligned}
\end{equation}
This design helps mitigate the inherent class imbalance: $\tau_{\text{benign}}$ is set higher to reduce the risk of majority-class dominance, while $\tau_{\text{fraud}}$ is kept relatively relaxed to maintain sufficient coverage of the minority fraud class. Using these confident subsets, we expand the labeled set for the next round:
\begin{equation}
  \mathcal{V}_{\text{pseudo}}^{(t)} = \hat{\mathcal{V}}_1 \cup \hat{\mathcal{V}}_0, \quad \mathcal{V}_L^{(t+1)} = \mathcal{V}_L^{(t)} \cup \mathcal{V}_{\text{pseudo}}^{(t)}.
\end{equation}
This iterative expansion progressively extends supervision coverage to more covert fraudsters across rounds, and is capped by the maximum number of self-training rounds.

Algorithm~\ref{alg:L2IR} summarizes the complete procedure of L2IR. The first stage (Lines 1--5) performs \textit{Behavior Intent Profiling} for all nodes. The second stage (Lines 6--14) conducts \textit{Connection Intent Reasoning} to identify and audit contradictory edges, producing the enriched features $H_v$. The third stage (Lines 15--21) runs the \textit{Adaptive Self-Training} loop. The final line returns the trained model, which can be used to detect camouflaged fraudulent nodes.

\subsection{Scalability Analysis}
\label{sec:scalability}
To ensure high scalability, L2IR decouples heavy LLM inference from iterative GNN training. LLM profiling and reasoning require $O(|\mathcal{V}| + |\mathcal{E}_{\text{suspicious}}|)$ forward passes, executed offline as one-time preprocessing. Selecting suspicious connections by node scores ($|\mathcal{E}_{\text{suspicious}}| \ll |\mathcal{E}_{\text{homo}}|$) avoids exhaustive evaluation over all edges. The self-training trains the GNN with $O(T \cdot |\mathcal{E}_{\text{homo}}| \cdot L \cdot d_h)$ complexity, where $L$ is the layer count and $d_h$ the hidden dimension. With rapid convergence ($T \leq 10$) and negligible feature fusion overhead, L2IR's computational cost remains comparable to standard GNN baselines, ensuring efficient scaling to large real-world graphs.

\section{Experiments}
In this subsection, we evaluate L2IR’s performance in two ways: (i) combining L2IR with CARE-GNN \cite{dou2020enhancing} as a standalone model, and (ii) applying L2IR as a plug-in to multiple GNN-based graph fraud detection models.

\begin{table}[t]
\centering
\small
\setlength{\tabcolsep}{3pt}
\renewcommand{\arraystretch}{1.15}
\caption{Statistics of the evaluation datasets.}
\label{tab:datasets}
\resizebox{\columnwidth}{!}{
\begin{tabular}{c c c c c c c}
\toprule
& \textbf{\#Nodes} 
& \begin{tabular}[c]{@{}c@{}}\textbf{\#Fraud}\\\textbf{(\%Ratio)}\end{tabular} 
& \textbf{Relation} 
& \textbf{\#Edges} 
& \begin{tabular}[c]{@{}c@{}}\textbf{Avg.}\\\textbf{Behav. Sim.}\end{tabular}
& \begin{tabular}[c]{@{}c@{}}\textbf{Avg.}\\\textbf{Conn. Sim.}\end{tabular} \\
\midrule
\multirow{4}{*}{\rotatebox{90}{\textbf{Amazon}}}
& \multirow{4}{*}{11,944} 
& \multirow{4}{*}{\shortstack{821\\(6.87\%)}} 
& \textit{U-P-U} & 177,547   & 0.71 & 16.72\% \\
& & & \textit{U-S-U} & 545,838 & 0.71 &  5.09\% \\
& & & \textit{U-V-U} & 3,036,733 & 0.70 &  6.35\% \\
& & & \textit{ALL}   & 3,515,011 & 0.70 &  7.93\% \\
\midrule
\multirow{5}{*}{\rotatebox{90}{\textbf{Yelp}}}
& \multirow{5}{*}{12,024} 
& \multirow{5}{*}{\shortstack{1,442\\(11.99\%)}} 
& \textit{U-P-U} & 877,840   & 0.78 &  9.80\% \\
& & & \textit{U-S-U} & 2,268,662 & 0.76 &  6.71\% \\
& & & \textit{U-V-U} & 1,174,921 & 0.74 & 15.45\% \\
& & & \textit{U-T-U} & 1,225,192 & 0.78 &  7.00\% \\
& & & \textit{ALL}   & 4,797,417 & 0.76 &  9.60\% \\
\bottomrule
\end{tabular}%
}
\end{table}

\subsection{Experimental Setup}

\textbf{Datasets.}
We evaluate L2IR on two real-world datasets, Amazon~\cite{mcauley2013amateurs} and Yelp~\cite{rayana2015collective}, which are widely used benchmarks for graph fraud detection. In both datasets, each node represents a user associated with statistical features derived from review metadata. Table \ref{tab:datasets} summarizes the statistics of the two datasets.

Specifically, the Amazon dataset comprises musical instrument reviews, structured into a graph where nodes represent users with 25-dimensional feature vectors. It contains three relation types~\cite{dou2020enhancing}: users reviewing at least one same product (U-P-U); users posting reviews with the same star rating within one week (U-S-U); and users with top-5\% mutual review text similarity (as measured by TF-IDF) (U-V-U). The Yelp dataset~\cite{rayana2015collective} consists of restaurant reviews from New York City, featuring 32-dimensional node attributes. Following similar relation construction strategies~\cite{dou2020enhancing}, it contains four relation types: users reviewing the same product within a sliding window of $K{=}10$ reviews (U-P-U); users assigning an identical extreme star rating within the same calendar month (U-S-U); users sharing top-$k$ ($k{=}100$) text similarity across all review texts (U-V-U); and users whose posting activity concentrates on the exact same calendar day regardless of the products reviewed, with co-occurrence capped at 50 (U-T-U). 

Avg. Behav. Sim. and Avg. Conn. Sim. in Table \ref{tab:datasets} reflect the camouflage degree of fraudulent nodes in the datasets, which will be elaborated in Section \ref{sec:camouflage}.

\textbf{Evaluation Metrics.} To comprehensively evaluate model performance, we employ three widely adopted metrics: AUROC, AUPRC and MacroF1. AUROC captures the model's ability to rank positive instances above negative ones across all possible thresholds, reflecting the overall separability between classes. MacroF1 averages the F1 scores of both positive and negative classes, providing a balanced evaluation that treats both classes equally regardless of class distribution. AUPRC focuses on performance on the positive class by balancing precision and recall, making it more sensitive to improvements on rare classes.

\begin{table*}[t] 
\centering 
\setlength{\tabcolsep}{4pt} 
\renewcommand{\arraystretch}{1.05} 
\caption{Performance of L2IR as a standalone model under various training ratios (AUROC, MacroF1, and AUPRC in percentage). \textbf{Bold}: best, \underline{Underline}: second best.} 
\label{tab:overall_full}

\resizebox{\textwidth}{!}{ 
\begin{tabular}{@{}c@{\hspace{7pt}}l@{\hspace{3pt}}c@{\hspace{5pt}}cccccccccc@{}}
\toprule 
\textbf{Dataset} & \textbf{Metric} & \textbf{Train\%} & \makecell[c]{Graph-\\SAGE}  & FdGars & \makecell[c]{Player-\\2Vec}  & \makecell[c]{CARE-\\GNN}  & \makecell[c]{Graph-\\Consis}  & BWGNN & PMP & \makecell[c]{Diff-\\Graph}  & RGTAN & \textbf{L2IR} \\ 
\midrule 
\multirow{12}{*}{{\textbf{Amazon}}} 
& \multirow{4}{*}{AUROC} 
  & 40\% & \std{75.24}{1.12} & \std{78.89}{0.94} & \std{81.24}{1.65} & \std{95.76}{0.31} & \std{96.15}{0.45} & \std{96.44}{1.63} & \stdb{97.78}{0.12} & \std{97.41}{0.21} & \std{96.01}{1.22} & \stdu{97.62}{0.28} \\ 
& & 30\% & \std{74.86}{0.94} & \std{78.49}{0.64} & \std{81.51}{1.68} & \std{95.40}{0.44} & \std{94.42}{1.14} & \std{94.97}{3.16} & \stdb{97.70}{0.11} & \std{96.97}{0.12} & \std{94.61}{2.17} & \stdu{97.58}{0.24} \\ 
& & 20\% & \std{74.92}{0.90} & \std{78.74}{0.53} & \std{82.03}{1.67} & \std{95.31}{0.50} & \std{94.03}{0.62} & \std{94.38}{2.00} & \stdu{97.44}{0.42} & \std{96.71}{0.51} & \std{93.38}{2.36} & \stdb{97.53}{0.27} \\ 
& & 10\% & \std{74.58}{0.73} & \std{78.47}{0.39} & \std{81.71}{1.63} & \std{94.97}{0.48} & \std{93.84}{0.56} & \std{93.29}{3.08} & \stdu{96.70}{0.46} & \std{96.03}{0.48} & \std{93.40}{0.93} & \stdb{97.10}{0.31} \\ 
\cmidrule{2-13} 
& \multirow{4}{*}{MacroF1} 
  & 40\% & \std{54.93}{2.70} & \std{64.96}{0.57} & \std{68.21}{1.80} & \std{88.84}{0.40} & \std{90.71}{0.34} & \std{91.51}{0.56} & \std{91.80}{0.56} & \std{90.40}{0.97} & \stdu{91.87}{0.42} & \stdb{92.46}{0.66} \\ 
& & 30\% & \std{57.45}{2.04} & \std{64.54}{0.51} & \std{68.42}{1.85} & \std{88.18}{0.65} & \std{90.48}{0.21} & \std{91.57}{0.40} & \stdb{91.84}{0.41} & \std{90.36}{0.74} & \std{91.23}{0.77} & \stdu{91.74}{0.53} \\ 
& & 20\% & \std{56.27}{2.64} & \std{64.91}{0.28} & \std{69.02}{1.80} & \std{88.54}{0.72} & \std{90.97}{0.18} & \std{91.19}{0.68} & \std{91.18}{0.35} & \std{90.20}{0.67} & \stdu{91.31}{0.66} & \stdb{91.68}{0.78} \\ 
& & 10\% & \std{54.88}{1.95} & \std{64.60}{0.62} & \std{68.71}{2.21} & \std{88.27}{1.05} & \stdu{90.53}{0.25} & \std{90.33}{0.41} & \std{90.30}{0.50} & \std{88.71}{1.22} & \std{89.56}{0.59} & \stdb{91.43}{1.24} \\ 
\cmidrule{2-13} 
& \multirow{4}{*}{AUPRC} 
  & 40\% & \std{20.67}{1.58} & \std{26.84}{0.98} & \std{40.25}{2.64} & \std{85.01}{0.65} & \std{84.48}{0.82} & \std{84.65}{2.81} & \stdu{88.74}{0.85} & \std{86.95}{0.95} & \std{87.70}{2.10} & \stdb{89.05}{0.61} \\ 
& & 30\% & \std{20.40}{1.08} & \std{26.41}{0.69} & \std{41.05}{2.97} & \std{84.28}{0.82} & \std{79.18}{5.63} & \std{82.23}{3.51} & \stdu{87.53}{0.40} & \std{86.24}{0.78} & \std{84.86}{2.51} & \stdb{87.81}{0.78} \\ 
& & 20\% & \std{19.81}{1.17} & \std{26.74}{0.31} & \std{41.71}{2.86} & \std{84.67}{0.95} & \std{82.36}{0.52} & \std{81.87}{2.67} & \stdu{86.67}{0.38} & \std{86.01}{0.63} & \std{84.66}{2.53} & \stdb{87.90}{0.84} \\ 
& & 10\% & \std{19.45}{0.37} & \std{26.60}{0.72} & \std{41.04}{3.91} & \std{83.87}{0.70} & \std{81.78}{0.58} & \std{81.30}{2.90} & \stdu{86.18}{0.37} & \std{84.19}{1.06} & \std{84.03}{1.91} & \stdb{87.52}{0.73} \\ 
\midrule 
\multirow{12}{*}{{\textbf{Yelp}}} 
& \multirow{4}{*}{AUROC} 
  & 40\% & \std{60.82}{1.38} & \std{72.80}{0.44} & \std{65.23}{1.51} & \std{88.62}{0.36} & \std{87.02}{0.39} & \std{88.26}{0.70} & \stdb{92.90}{0.25} & \std{88.11}{0.25} & \std{89.35}{0.86} & \stdu{91.85}{0.48} \\ 
& & 30\% & \std{61.57}{1.56} & \std{73.12}{0.17} & \std{65.23}{1.91} & \std{87.95}{0.69} & \std{86.21}{0.35} & \std{88.69}{0.49} & \stdu{91.38}{0.20} & \std{87.48}{0.41} & \std{88.02}{0.68} & \stdb{91.72}{0.62} \\ 
& & 20\% & \std{61.02}{1.34} & \std{73.21}{0.25} & \std{63.67}{2.81} & \std{87.40}{0.62} & \std{84.89}{2.18} & \std{87.56}{1.09} & \stdu{88.50}{0.31} & \std{86.92}{0.71} & \std{86.97}{0.88} & \stdb{90.42}{0.77} \\ 
& & 10\% & \std{60.02}{1.61} & \std{73.03}{0.34} & \std{61.08}{9.32} & \stdu{86.33}{0.45} & \std{81.64}{3.21} & \std{85.87}{0.54} & \std{86.00}{0.17} & \std{85.94}{0.76} & \std{86.05}{0.59} & \stdb{89.23}{0.49} \\ 
\cmidrule{2-13} 
& \multirow{4}{*}{MacroF1} 
  & 40\% & \std{48.47}{2.17} & \std{60.49}{0.41} & \std{59.52}{1.32} & \std{73.78}{1.95} & \std{75.97}{0.01} & \std{78.27}{0.62} & \stdb{78.92}{0.66} & \std{76.25}{0.66} & \stdu{78.41}{1.10} & \std{78.15}{1.62} \\ 
& & 30\% & \std{49.60}{1.90} & \std{60.68}{0.31} & \std{59.57}{1.74} & \std{72.99}{0.21} & \std{74.56}{0.32} & \stdb{78.39}{0.85} & \std{77.89}{0.17} & \std{75.89}{0.78} & \std{77.96}{0.65} & \stdu{78.02}{0.54} \\ 
& & 20\% & \std{49.38}{2.25} & \std{60.74}{0.34} & \std{58.23}{2.59} & \std{71.94}{1.07} & \std{74.17}{1.89} & \std{75.57}{0.59} & \std{74.83}{0.38} & \std{74.99}{0.94} & \stdu{75.88}{1.06} & \stdb{77.45}{1.56} \\ 
& & 10\% & \std{49.72}{2.69} & \std{60.72}{0.28} & \std{57.40}{4.07} & \std{70.93}{1.75} & \std{71.51}{2.68} & \std{74.11}{1.42} & \std{72.10}{0.37} & \std{73.21}{1.20} & \stdu{74.67}{1.18} & \stdb{75.17}{1.48} \\ 
\cmidrule{2-13} 
& \multirow{4}{*}{AUPRC} 
  & 40\% & \std{19.37}{1.45} & \std{24.45}{0.68} & \std{24.04}{2.77} & \std{64.81}{0.96} & \std{57.93}{0.88} & \std{66.58}{1.18} & \stdb{71.01}{0.98} & \std{62.70}{0.86} & \std{68.87}{1.74} & \stdu{69.86}{0.89} \\ 
& & 30\% & \std{19.82}{1.41} & \std{24.72}{0.19} & \std{24.57}{2.81} & \std{62.31}{1.07} & \std{52.78}{1.66} & \std{67.19}{2.51} & \stdu{67.58}{0.69} & \std{62.32}{1.62} & \std{66.36}{1.62} & \stdb{68.51}{1.19} \\ 
& & 20\% & \std{18.98}{1.29} & \std{24.99}{0.39} & \std{22.28}{4.40} & \std{60.57}{1.37} & \std{50.59}{6.24} & \std{64.23}{1.54} & \std{60.29}{1.19} & \std{60.00}{1.80} & \stdu{64.39}{1.36} & \stdb{65.44}{1.06} \\ 
& & 10\% & \std{18.79}{1.41} & \std{24.90}{0.55} & \std{21.44}{5.82} & \std{57.74}{1.98} & \std{43.15}{9.99} & \std{58.74}{2.18} & \std{53.93}{0.25} & \std{56.87}{1.87} & \stdu{58.82}{1.75} & \stdb{60.93}{2.03} \\ 
\bottomrule 
\end{tabular} 
} 
\end{table*}

\textbf{Baselines.} 
We compare L2IR with nine state-of-the-art baselines for graph fraud detection, including GraphSAGE~\cite{hamilton2017inductive}, FdGars~\cite{wang2019fdgars}, Player2Vec~\cite{zhang2019key}, CARE-GNN~\cite{dou2020enhancing}, GraphConsis~\cite{liu2020alleviating}, BWGNN~\cite{tang2022rethinking}, PMP~\cite{zhuo2024partitioning}, DiffGraph~\cite{li2025diffgraph}, and RGTAN~\cite{xiang2025enhancing}. We reproduce all baselines using the publicly released codebases provided by the respective authors.

\textbf{Implementation.} For a fair comparison, each baseline is configured according to the hyperparameter settings recommended in its original paper, and all methods are optimized with Adam. Besides, we set the training batch size to 128 on Amazon and 256 on Yelp. The confidence thresholds are set as $\tau_{\text{fraud}}=0.90$ and $\tau_{\text{benign}}=0.95$ on Amazon, and $\tau_{\text{fraud}}=0.80$ and $\tau_{\text{benign}}=0.85$ on Yelp. The maximum number of self-training rounds is set to 3 on Amazon and 2 on Yelp. For behavior intent profiling, we set the number of retrieved exemplars per class to $k=2$. For connection intent reasoning, we set $\tau_h=0.80$, $\tau_l=0.20$, and $s=4000$ to retain suspicious connections ranked by contradictory magnitude while bounding the LLM inference cost. Each method undergoes ten trials with different random seeds, and we report the mean and standard deviation. For the LLM components in L2IR, we use \textit{Llama-3.1-8B}~\cite{grattafiori2024Llama} as the default backbone, and additionally evaluate \textit{Qwen3-8B}~\cite{yang2025qwen3} when L2IR is applied as a plug-in to multiple GNN-based methods. All experiments are executed on NVIDIA RTX 3090 24GB GPU.

\subsection{Fraudster Camouflage Analysis}
\label{sec:camouflage}
To verify the degree of fraudster camouflage in our evaluation datasets, we employ two complementary metrics introduced in~\cite{liu2020alleviating}. The first is \textit{behavior similarity} (abbreviated as Behav.\ Sim.), which ranges from 0 to 1 and quantifies the extent to which a fraudulent node's behavior profile resembles those of its neighbors. It is computed as the average pairwise Gaussian kernel (RBF) similarity between the feature vector of a fraudulent node and those of its connected neighbors. A value closer to 1 indicates stronger behavioral resemblance, making fraudulent nodes more difficult to distinguish from benign ones. The second is \textit{connection similarity} (abbreviated as Conn.\ Sim.), defined as the fraction of a fraudulent node's neighbors that are also fraudulent. It directly reflects the degree to which fraudsters embed themselves within benign neighborhoods: a lower value indicates deeper camouflage, as the fraudulent node is surrounded predominantly by benign neighbors.
\begin{figure}[htbp]
  \centering
  \includegraphics[width=\columnwidth]{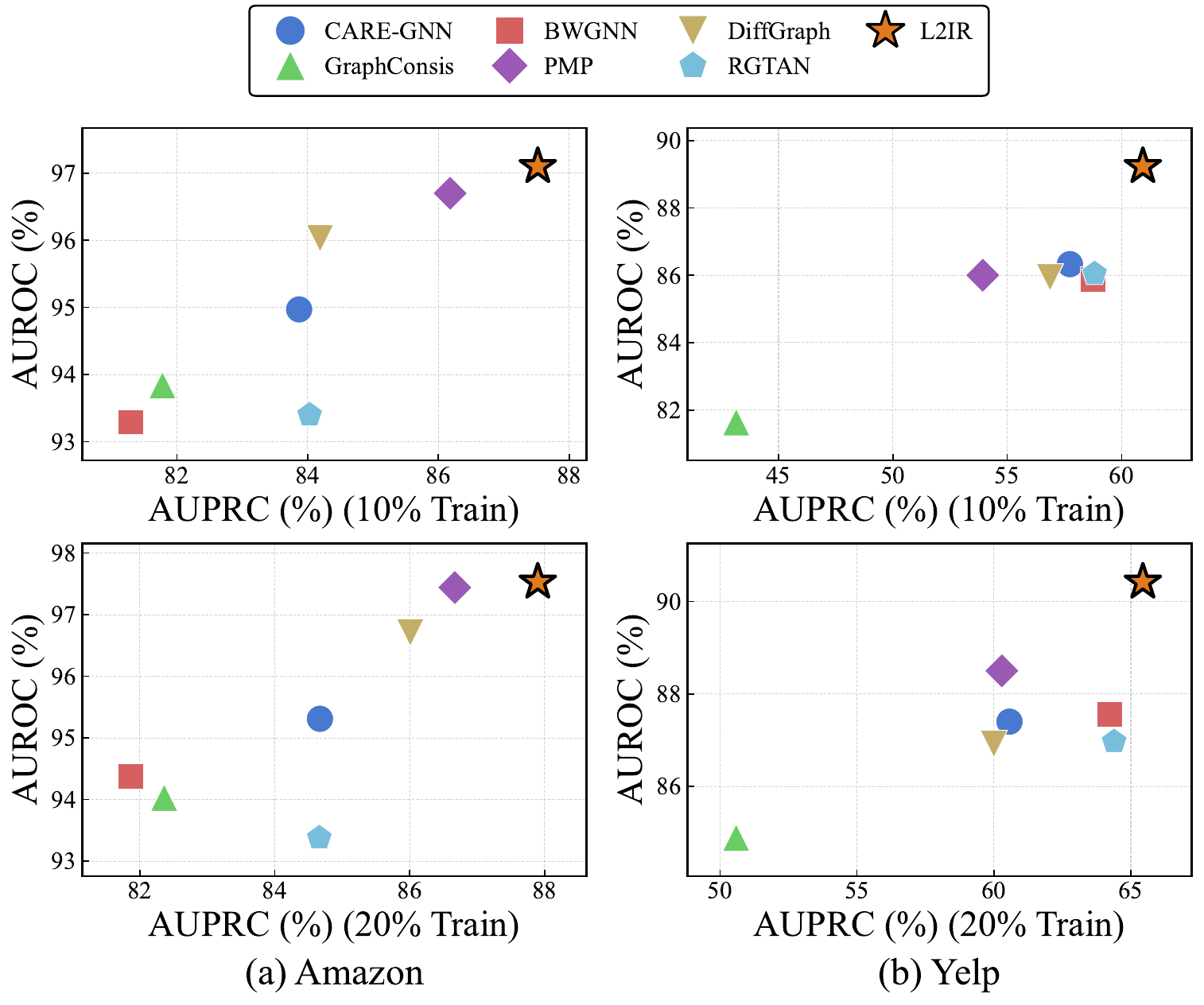}
  \caption{AUROC vs.\ AUPRC at 10\% and 20\% training ratios. Each point represents a method; the upper-right corner indicates better joint performance.}
  \label{fig:fraud_scatter_0.1+0.2}
\end{figure}
Table~\ref{tab:datasets} reports both metrics across all relation types on the two datasets. From the table, the behavior similarity scores of both datasets are consistently high, maintaining an average of 0.70 or higher across all relations. Connection similarity scores are consistently low, with the majority of each fraudulent node's connections leading to benign neighbors. On Yelp, for instance, only 6.71\% and 7.00\% of the neighbors of fraudulent nodes under U-S-U and U-T-U relations are also fraudsters. These results demonstrate our evaluation datasets exhibit a high degree of camouflaged fraud.

\begin{table}[t]
\centering
\small
\setlength{\tabcolsep}{5pt}
\renewcommand{\arraystretch}{1.05}
\caption{Performance of L2IR as a plug-in module to GNN-based methods under the 50\% labeled setting with a fixed 40\% train ratio (AUROC and AUPRC in percentage).}
\label{tab:overall_limited}
\begin{tabular}{l cc cc}
\toprule
\multicolumn{1}{c}{\multirow{2}{*}{\textbf{Methods}}}
  & \multicolumn{2}{c}{\textbf{Amazon}}
  & \multicolumn{2}{c}{\textbf{Yelp}} \\
\cmidrule(l{4pt}r{4pt}){2-3} \cmidrule(l{4pt}){4-5}
  & \textbf{AUROC} & \textbf{AUPRC}
  & \textbf{AUROC} & \textbf{AUPRC} \\
\midrule
GraphSAGE                            & \std{73.99}{1.42}          & \std{25.23}{1.07}          & \std{61.10}{1.32}          & \std{19.84}{0.86} \\
~~+L2IR$_{\textit{Qwen3-8B}}$        & \std{75.05}{0.73}          & \std{26.12}{0.97}          & \std{63.84}{1.28}         & \std{21.61}{1.82} \\
~~+L2IR$_{\textit{Llama-3.1-8B}}$    & \std{75.19}{1.68}         & \std{26.24}{1.43}         & \std{63.78}{1.35}          & \std{21.03}{1.71} \\
\midrule

FdGars                               & \std{80.73}{0.72}          & \std{32.61}{0.92}          & \std{74.40}{1.24}          & \std{26.02}{1.81} \\
~~+L2IR$_{\textit{Qwen3-8B}}$        & \std{81.68}{0.49}         & \std{33.34}{0.76}          & \std{75.62}{1.13}          & \std{26.63}{1.54} \\
~~+L2IR$_{\textit{Llama-3.1-8B}}$    & \std{81.52}{0.53}          & \std{33.45}{0.82}         & \std{75.81}{1.09}         & \std{26.80}{1.62} \\
\midrule

Player2Vec                           & \std{82.92}{1.23}          & \std{46.51}{2.55}          & \std{66.32}{0.61}          & \std{26.90}{1.92} \\
~~+L2IR$_{\textit{Qwen3-8B}}$        & \std{85.85}{0.64}          & \std{49.98}{1.71}         & \std{68.97}{1.02}          & \std{29.96}{1.01} \\
~~+L2IR$_{\textit{Llama-3.1-8B}}$    & \std{85.93}{0.71}         & \std{48.81}{0.83}          & \std{69.14}{1.43}         & \std{30.11}{0.89} \\
\midrule

GraphConsis                          & \std{92.46}{0.31}          & \std{80.51}{0.66}          & \std{84.03}{3.51}          & \std{49.48}{6.02} \\
~~+L2IR$_{\textit{Qwen3-8B}}$        & \std{94.33}{1.16}         & \std{84.37}{0.85}         & \std{84.96}{1.10}          & \std{51.72}{4.65} \\
~~+L2IR$_{\textit{Llama-3.1-8B}}$    & \std{94.21}{1.24}          & \std{84.19}{0.42}          & \std{85.12}{1.18}         & \std{51.07}{3.43} \\
\midrule

CARE-GNN                             & \std{90.60}{0.56}          & \std{79.63}{1.06}          & \std{83.11}{0.74}          & \std{58.22}{2.35} \\
~~+L2IR$_{\textit{Qwen3-8B}}$        & \std{94.19}{0.74}         & \std{84.36}{0.98}          & \std{87.70}{0.67}          & \std{65.01}{1.61} \\
~~+L2IR$_{\textit{Llama-3.1-8B}}$    & \std{93.82}{0.81}          & \std{84.93}{0.89}         & \std{88.04}{0.91}         & \std{64.82}{1.87} \\
\midrule

BWGNN                                & \std{86.81}{0.67}          & \std{84.99}{0.70}          & \std{84.85}{0.26}          & \std{54.43}{3.53} \\
~~+L2IR$_{\textit{Qwen3-8B}}$        & \std{96.73}{0.39}         & \std{85.96}{0.83}         & \std{87.47}{1.75}         & \std{68.02}{3.13} \\
~~+L2IR$_{\textit{Llama-3.1-8B}}$    & \std{96.59}{0.47}          & \std{85.21}{0.95}          & \std{86.32}{1.58}          & \std{66.96}{2.84} \\
\midrule

PMP                                  & \std{88.01}{0.37}          & \std{86.56}{0.30}          & \std{90.76}{0.24}          & \std{73.12}{1.19} \\
~~+L2IR$_{\textit{Qwen3-8B}}$        & \std{94.72}{0.46}          & \std{88.20}{1.17}         & \std{91.64}{0.49}         & \std{78.76}{0.92} \\
~~+L2IR$_{\textit{Llama-3.1-8B}}$    & \std{95.20}{0.79}         & \std{87.64}{0.55}          & \std{91.07}{0.36}          & \std{78.25}{0.61} \\
\midrule

DiffGraph                            & \std{96.20}{0.39}          & \std{83.39}{2.08}          & \std{85.98}{0.70}          & \std{56.09}{3.24} \\
~~+L2IR$_{\textit{Qwen3-8B}}$        & \std{97.05}{0.25}         & \std{88.22}{0.62}         & \std{87.68}{0.84}          & \std{63.98}{2.50} \\
~~+L2IR$_{\textit{Llama-3.1-8B}}$    & \std{96.91}{0.28}          & \std{88.14}{0.57}          & \std{87.77}{0.41}         & \std{63.01}{2.18} \\
\midrule
RGTAN                                & \std{93.65}{2.27}          & \std{82.48}{3.58}          & \std{84.33}{2.86}          & \std{65.45}{3.45} \\
~~+L2IR$_{\textit{Qwen3-8B}}$        & \std{94.51}{1.74}           & \std{84.53}{2.17}           & \std{89.02}{2.16}           & \std{73.72}{2.39} \\
~~+L2IR$_{\textit{Llama-3.1-8B}}$    & \std{94.28}{1.63}           & \std{84.66}{2.08}           & \std{88.91}{2.21}           & \std{73.14}{2.53} \\
\bottomrule
\end{tabular}
\end{table}

\subsection{Overall Performance}

\textbf{L2IR as a Standalone Model.}
Table~\ref{tab:overall_full} presents the accuracy comparison between L2IR combined with CARE-GNN \cite{dou2020enhancing} as a standalone model and the baseline methods. We adopt CARE-GNN as the fixed backbone due to its high computational efficiency. All experiments are conducted on fully annotated datasets under training ratios ranging from 10\% to 40\%, with the remaining nodes used as the test set.

As shown in Table~\ref{tab:overall_full}, L2IR achieves the best or second-best results under most scenarios, and the advantage tends to become more pronounced as the training ratio decreases. Notably, with only 20\% and 10\% training data, L2IR outperforms all baselines across all metrics and consistently achieves the best results on both datasets. Specifically, on Amazon, L2IR achieves the best AUPRC at both 20\% and 10\% training ratios, surpassing the best baseline (PMP) by relative margins of 1.42\% and 1.55\%, respectively. On Yelp, at 20\% training ratio, L2IR improves AUROC by 2.17\% over PMP and AUPRC by 1.63\% over RGTAN; at 10\%, the relative gains further increase to 3.36\% in AUROC over CARE-GNN and 3.59\% in AUPRC over RGTAN. For MacroF1, L2IR achieves improvements of 2.07\% over RGTAN on Yelp (20\%) and 0.99\% over GraphConsis on Amazon (10\%). As further illustrated in Figure~\ref{fig:fraud_scatter_0.1+0.2}, comparing AUROC against AUPRC at these low training ratios shows that L2IR consistently occupies the optimal upper-right region. These results confirm that by inferring connection intent to expose camouflaged fraud, L2IR improves detecting camouflaged fraudsters while maintaining stable fraud-benign discrimination, and becomes especially effective under limited supervision.

\begin{table}[t]
\centering
\small
\setlength{\tabcolsep}{4pt}
\renewcommand{\arraystretch}{1.05}
\caption{Ablation study results under 50\% labeled setting with a fixed 40\% train ratio (AUROC and AUPRC in percentage).}
\label{tab:ablation}
\resizebox{\columnwidth}{!}{%
\begin{tabular}{c ccc cc cc}
\toprule
\multirow{2}{*}{\textbf{Methods}}
  & \multicolumn{3}{c}{\textbf{Variant}}
  & \multicolumn{2}{c}{\textbf{Amazon}}
  & \multicolumn{2}{c}{\textbf{Yelp}} \\
\cmidrule(l{4pt}r{4pt}){2-4}
\cmidrule(l{4pt}r{4pt}){5-6}
\cmidrule(l{4pt}){7-8}
  & \textbf{BI} & \textbf{CI} & \textbf{ST}
  & \textbf{AUROC} & \textbf{AUPRC}
  & \textbf{AUROC} & \textbf{AUPRC} \\
\midrule
\multirow{4}{*}{BWGNN}
  & \checkmark & \checkmark & \checkmark
    & \std{96.73}{0.39} & \std{85.96}{0.83}
    & \std{87.47}{1.75} & \std{68.02}{3.13} \\
  &            & \checkmark & \checkmark
    & \std{93.84}{0.57} & \std{85.62}{0.85}
    & \std{86.54}{1.40} & \std{63.29}{1.17} \\
  & \checkmark &            & \checkmark
    & \std{90.37}{0.98} & \std{85.14}{1.25}
    & \std{85.77}{1.85} & \std{60.19}{2.27} \\
  & \checkmark & \checkmark &
    & \std{92.89}{0.48} & \std{85.47}{2.19}
    & \std{86.21}{1.36} & \std{64.81}{0.99} \\
\midrule
\multirow{4}{*}{PMP}
  & \checkmark & \checkmark & \checkmark
    & \std{94.72}{0.46} & \std{88.20}{1.17}
    & \std{91.64}{0.49} & \std{78.76}{0.92} \\
  &            & \checkmark & \checkmark
    & \std{93.28}{0.90} & \std{87.23}{0.73}
    & \std{91.22}{0.51} & \std{76.87}{0.73} \\
  & \checkmark &            & \checkmark
    & \std{91.35}{0.61} & \std{86.91}{1.04}
    & \std{91.07}{0.74} & \std{75.05}{0.97} \\
  & \checkmark & \checkmark &
    & \std{92.85}{0.54} & \std{87.52}{0.87}
    & \std{91.51}{0.62} & \std{76.50}{0.79} \\
\midrule
\multirow{4}{*}{DiffGraph}
  & \checkmark & \checkmark & \checkmark
    & \std{97.05}{0.25} & \std{88.22}{0.62}
    & \std{87.68}{0.84} & \std{63.98}{2.50} \\
  &            & \checkmark & \checkmark
    & \std{96.88}{0.37} & \std{86.86}{0.90}
    & \std{87.10}{0.85} & \std{61.81}{2.09} \\
  & \checkmark &            & \checkmark
    & \std{96.43}{0.63}  & \std{85.13}{0.75}
    & \std{86.21}{1.04}  & \std{59.48}{2.36} \\
  & \checkmark & \checkmark &
    & \std{96.71}{0.42}  & \std{86.69}{1.16}
    & \std{86.72}{0.96}  & \std{62.04}{2.14} \\
\midrule
\multirow{4}{*}{RGTAN}
  & \checkmark & \checkmark & \checkmark
    & \std{94.51}{1.74} & \std{84.53}{2.17}
    & \std{89.02}{2.16} & \std{73.72}{2.39} \\
  &            & \checkmark & \checkmark
    & \std{94.33}{1.24} & \std{84.01}{2.01}
    & \std{87.35}{1.01} & \std{70.66}{1.65} \\
  & \checkmark &            & \checkmark
    & \std{93.94}{1.78} & \std{82.82}{2.14}
    & \std{86.84}{2.17} & \std{68.47}{2.37} \\
  & \checkmark & \checkmark &
    & \std{94.37}{1.39} & \std{83.69}{1.46}
    & \std{87.79}{2.32} & \std{71.37}{2.49} \\
\bottomrule
\end{tabular}%
}
\end{table}

\textbf{L2IR as a Plug-in for GNNs.}
Table~\ref{tab:overall_limited} reports the results of L2IR as a plug-in module for improving GNN-based fraud detection models. The experiments are conducted under more stringent conditions to evaluate the robustness of L2IR: 50\% of the fraud labels are removed, and the training ratio is set to 40\%. L2IR is instantiated with two base LLMs, \textit{Qwen3-8B} and \textit{Llama-3.1-8B}, for performance comparison.

As shown in the table, L2IR consistently improves all GNN-based methods on both datasets across the two LLM variants. With \textit{Qwen3-8B}, L2IR brings average gains of 3.19\% in AUROC and 2.57\% in AUPRC on Amazon, and 2.45\% and 5.54\% on Yelp, respectively. With \textit{Llama-3.1-8B}, L2IR obtains similar average gains of 3.14\% and 2.37\% on Amazon, and 2.34\% and 5.07\% on Yelp. The gains also hold for strong GNN-based methods such as RGTAN and DiffGraph. For example, L2IR improves RGTAN by 8.27\% in AUPRC on Yelp and DiffGraph by 4.83\% in AUPRC on Amazon. These results confirm that L2IR generalizes well to different GNN-based methods, where semantic evidence derived from behavior and connection intents jointly provides reliable signals of camouflaged fraud, and the self-training mechanism effectively compensates for the insufficient labels. The consistent gains across the two LLM variants also suggest that L2IR is not sensitive to the choice of the backbone language model.

\subsection{Ablation Study}
To validate the contribution of each key module in L2IR, we conduct ablation studies by integrating L2IR with five representative GNN models. Specifically, we ablate three core modules—\textit{Behavior Intent Profiling} (BI), \textit{Connection Intent Reasoning} (CI), and \textit{Adaptive Self-Training} (ST). For these ablation studies, we adopt the same setting used in Table~\ref{tab:overall_limited} where 50\% of the fraud labels are removed.

Table~\ref{tab:ablation} presents the ablation results of the three core components on both datasets. As shown in the table, removing any module consistently degrades performance across different GNN-based methods, confirming that these modules provide complementary benefits. Among them, CI contributes the most, as excluding it leads to the largest drop in most metrics, particularly in AUPRC. For example, AUPRC drops by 7.83\%, 5.25\%, and 4.50\% on Yelp with BWGNN, RGTAN, and DiffGraph, respectively, confirming that \textit{Connection Intent Reasoning} is crucial for improving detection reliability under camouflaged fraud. Removing BI also reduces performance noticeably, with AUPRC drops of 4.73\% and 3.06\% on Yelp with BWGNN and RGTAN, respectively, showing that semantic evidence from \textit{Behavior Intent Profiling} helps retain fraud signals before graph aggregation. Excluding ST further decreases AUPRC by 3.21\%, 2.26\%, and 2.35\% on Yelp with BWGNN, PMP, and RGTAN, respectively, demonstrating the effectiveness of expanding supervision with \textit{Adaptive Self-Training}.

\setlength{\abovecaptionskip}{6pt}


\begin{figure}[t]
\centering
\includegraphics[width=\linewidth]{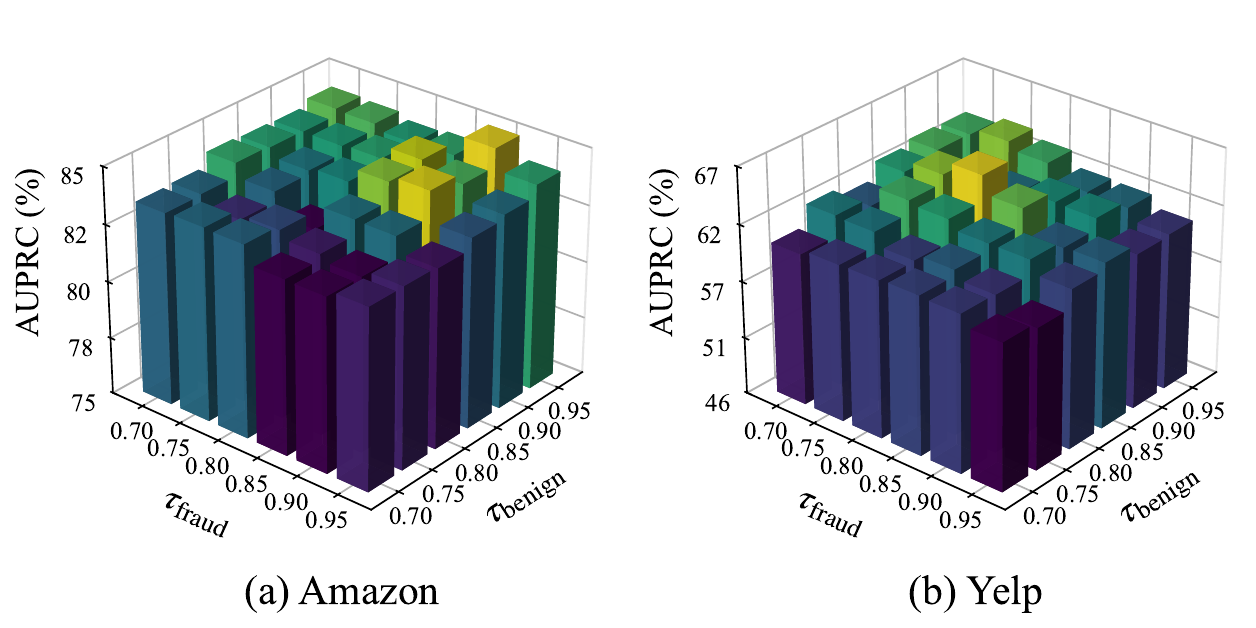}
\caption{Parameter sensitivity analysis on confidence thresholds $\tau_{\text{fraud}}$ and $\tau_{\text{benign}}$.}
\label{fig:sensitivity_threshold}
\end{figure}

\subsection{Parameter Sensitivity Analysis}
Using CARE-GNN as the backbone~\cite{dou2020enhancing}, we investigate the sensitivity of L2IR to three key parameter types on both datasets: (i) the confidence thresholds $\tau_{\text{fraud}}$ and $\tau_{\text{benign}}$ for adaptive self-training, (ii) the class balance ratio (positive:negative) of the training set, and (iii) the maximum number of self-training rounds.

Figure~\ref{fig:sensitivity_threshold} presents the impact of the confidence thresholds $\tau_{\text{fraud}}$ and $\tau_{\text{benign}}$ on AUPRC. Amazon exhibits a broad stable region across most configurations, with peak AUPRC of 84.92\% when $\tau_{\text{benign}}=0.95$ and $\tau_{\text{fraud}}=0.90$, suggesting relatively low sensitivity to the exact threshold values. Yelp shows greater sensitivity, with AUPRC peaking at 65.03\% when $\tau_{\text{benign}}=0.85$ and $\tau_{\text{fraud}}=0.80$, and dropping when $\tau_{\text{fraud}}$ is set too high relative to $\tau_{\text{benign}}$, as a looser $\tau_{\text{benign}}$ admits more low-confidence benign pseudo-labels into training, diluting fraud signals and shifting the model toward the majority class.

\begin{figure}[t]
\centering
\includegraphics[width=\columnwidth]{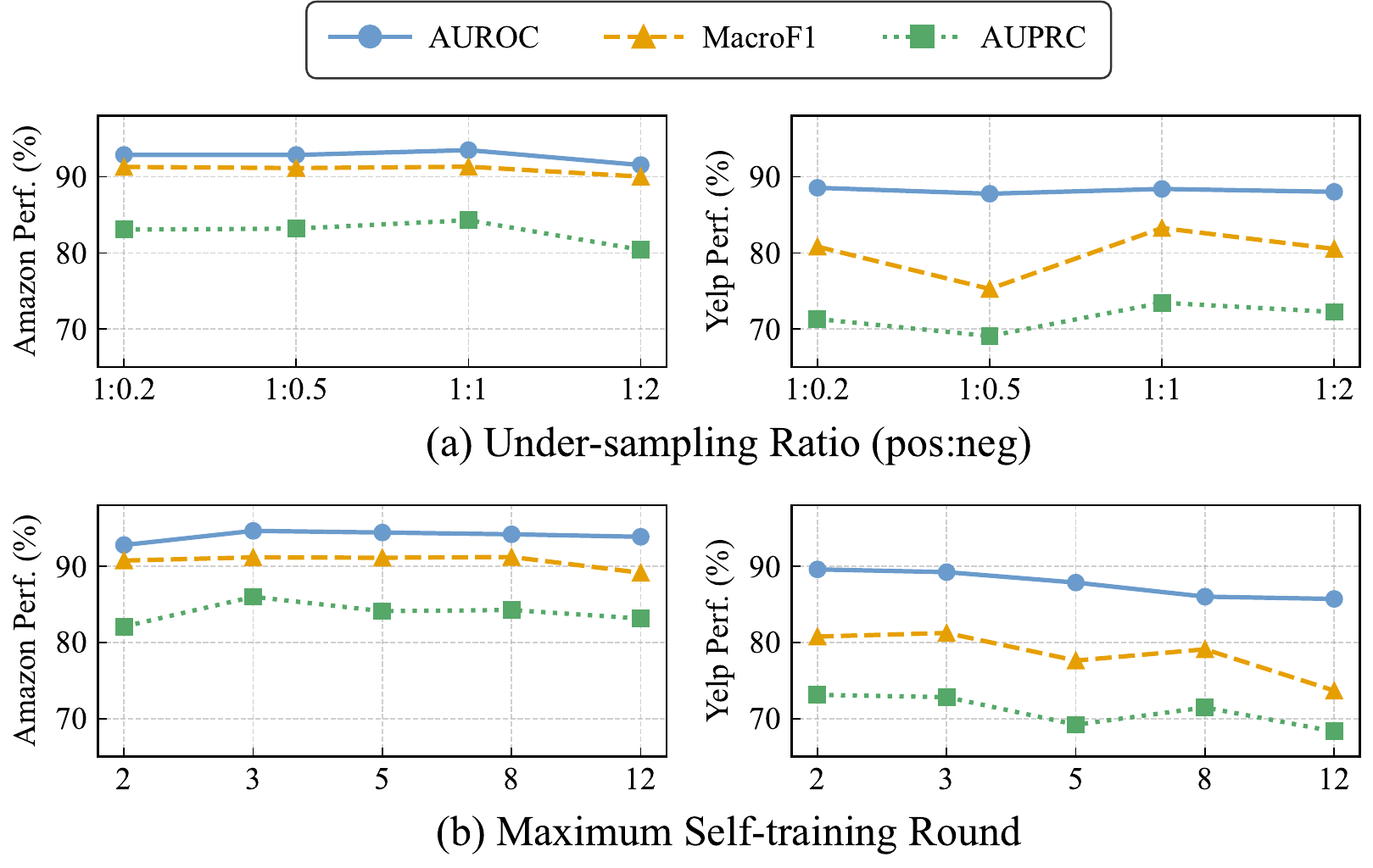}
\caption{Parameter sensitivity analysis on under-sampling ratio and maximum self-training round.}
\label{fig:sensitivity_line}
\end{figure}

\begin{figure}[t]
\centering
\includegraphics[width=\columnwidth]{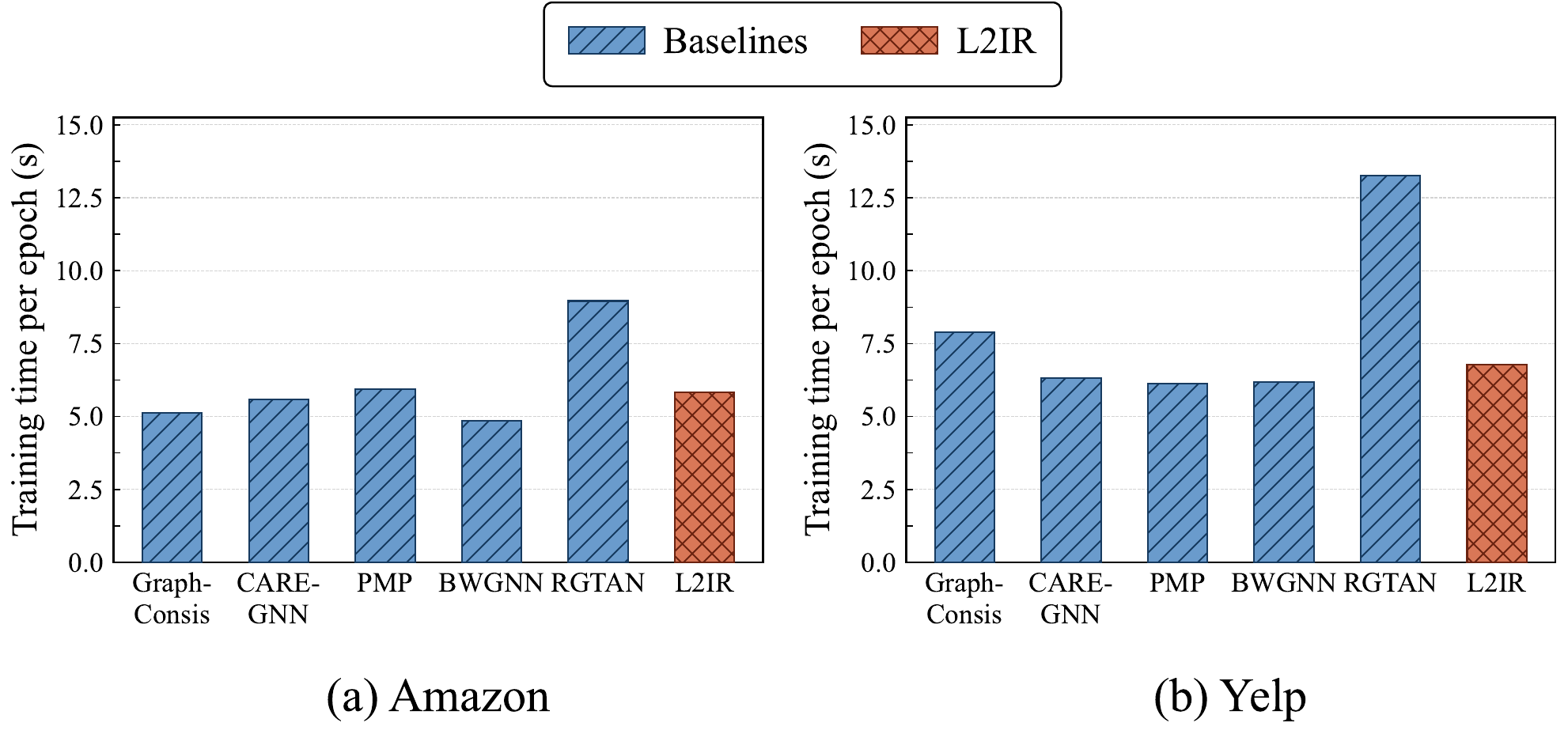}
\caption{Efficiency comparison on two datasets.}
\label{fig:efficiency_bar}
\end{figure}

Figure~\ref{fig:sensitivity_line}(a) evaluates the impact of varying the class balance ratio. We train L2IR under different balance ratios. Performance peaks at a 1:1 ratio on both datasets. The decline at the 1:2 ratio, particularly in MacroF1 and AUPRC, indicates that an excess of negative samples shifts the learning bias toward the benign majority. Overall, L2IR exhibits strong robustness to the class balance ratio in the training set.

Figure~\ref{fig:sensitivity_line}(b) illustrates the influence of maximum self-training rounds. Optimal results emerge at early stages (round 3 for Amazon and round 2 for Yelp). Extending the process to additional rounds may degrade performance, as noisy pseudo-labels can accumulate in the training set.

\subsection{Efficiency Analysis}
We evaluate the computational efficiency of L2IR by comparing its per-epoch training time against representative baselines. As shown in Figure~\ref{fig:efficiency_bar}, on Amazon, L2IR achieves a per-epoch training time of 5.83s, which is comparable to CARE-GNN at 5.57s and PMP at 5.95s, while being significantly faster than RGTAN at 8.96s. 
On Yelp, L2IR requires 6.77s per epoch, remaining within the range of standard baselines and taking nearly half the time of RGTAN at 13.26s, where RGTAN incurs additional cost from temporal attention and risk-aware neighborhood modeling. 
This efficiency of L2IR is achieved by fully decoupling the LLM inference for both behavior intent profiling and connection intent reasoning as a one-time offline preprocessing step, which bounds the online training cost to standard GNN complexity.

\section{Conclusions}
In this paper, we propose L2IR, an LLM-driven latent intent
revealing framework for graph fraud detection in the presence of widespread camouflage. We design behavior intent profiling and connection intent reasoning to model the intents behind node interactions. By leveraging LLMs to infer hidden interaction intents, L2IR enhances the distinction between camouflaged fraudsters and benign entities. We also introduce adaptive self-training to improve training performance under limited fraud labels. Experiments on two real-world benchmarks show that L2IR consistently outperforms state-of-the-art baselines, particularly under low training ratios and with scarce labeled fraud. Ablation studies verify the effectiveness of the proposed designs, and importantly, these designs introduce no significant overhead compared to standard GNNs, as LLM inference is decoupled as a one-time offline preprocessing step.

\FloatBarrier
\appendix

\section{Prompt Templates}
\label{app:complete_prompt}

\begin{prompttemplate}{Prompt template for Behavior Intent Profiling}
(*@\textbf{\texttt{SYSTEM\_PROMPT}}@*) = """
You are a senior fraud detection analyst specializing in review behavior and relation camouflage. Your task is to infer behavior intent. Strictly follow all provided constraints and requirements.
"""

(*@\textbf{\texttt{USER\_PROMPT}}@*) = """
You are given reference cases, target user information, graph context, neighbor distribution, and review history. The Amazon dataset contains e-commerce product reviews, where users may show fraudulent behavior through abnormal review activity and camouflaged relations.

Few-Shot Exemplars: (*@\textcolor{placeholderblue}{\{few-shot\_exemplars\}}@*)
Target User: (*@\textcolor{placeholderblue}{\{target\_user\}}@*)
Target Statistics: (*@\textcolor{placeholderblue}{\{target\_statistics\}}@*)
Graph Relation Context: (*@\textcolor{placeholderblue}{\{graph\_relation\_context\}}@*)
Review Traces: (*@\textcolor{placeholderblue}{\{review\_traces\}}@*)

Task:
Analyze the target user's behavior intent. Focus on activity pattern, rating behavior, review content, helpfulness, graph context, and possible fraud signals.

Requirements:
1) **Information source**: Use only the provided information.
2) **Reference use**: Use reference cases only for comparison. Do not treat them as target labels.
3) **Context use**: Use graph relation context as supporting information. Do not infer the target label from neighbor labels alone.
4) **Balanced signals**: Consider both fraud signals and benign signals.
5) **Intent grounding**: Separate observed behavior from inferred behavior intent.
6) **Output format**: Return only four sections: User Profile Summary, Behavior Pattern Analysis, Fraud Signal Analysis, and Overall Assessment. Do not output the final class label. The output should read naturally and contain no phrasing that suggests it was produced by a language model.
"""
\end{prompttemplate}

\begin{prompttemplate}{Prompt template for Connection Intent Reasoning}
(*@\textbf{\texttt{SYSTEM\_PROMPT}}@*) = """
You are a senior fraud audit analyst specializing in review graphs and relation camouflage. Your task is to analyze the intent behind a suspicious connection between a suspected fraudster and a likely benign user, and assess whether this connection provides supportive or misleading evidence for fraud detection. Strictly follow all provided constraints and requirements.
"""

(*@\textbf{\texttt{USER\_PROMPT}}@*) = """
You are given a suspicious connection, preliminary risk information, and the review histories of two connected users.
The Amazon dataset contains e-commerce product reviews, where suspicious user connections may either support fraud detection or reflect relation camouflage.

Target Connection: (*@\textcolor{placeholderblue}{\{connection\_metadata\}}@*)
User A Review Traces: (*@\textcolor{placeholderblue}{\{user\_a\_reviews\}}@*)
User B Review Traces: (*@\textcolor{placeholderblue}{\{user\_b\_reviews\}}@*)

Task:
Analyze the connection between User A and User B. Focus on behavior difference, shared products, rating pattern, review timing, review content, helpfulness, and connection intent. Use only connection related information and do not repeat full user profiles.

Requirements:
1) **Information source**: Use only the provided information.
2) **Role use**: Use preliminary roles and scores only as audit signals. Do not treat them as ground truth labels.
3) **Balanced judgment**: Consider both supportive and misleading interpretations.
4) **Intent grounding**: Separate observed behavior from inferred connection intent.
5) **Verdict format**: In Risk Verdict, return the risk level as Low, Medium, or High, confidence as a number between 0 and 1, and exactly three key signals.
6) **Output format**: Return only five sections: Connection Overview, Behavior Difference, Connection Intent Analysis, Counter Evidence and Uncertainty, and Risk Verdict. The output should read naturally and contain no phrasing that suggests it was produced by a language model.
"""
\end{prompttemplate}

\bibliographystyle{ACM-Reference-Format}
\bibliography{references}

@article{akoglu2015graph,
  title={Graph based anomaly detection and description: a survey},
  author={Akoglu, Leman and Tong, Hanghang and Koutra, Danai},
  journal={Data mining and knowledge discovery},
  volume={29},
  number={3},
  pages={626--688},
  year={2015},
  publisher={Springer}
}

@inproceedings{rayana2015collective,
  title={Collective opinion spam detection: Bridging review networks and metadata},
  author={Rayana, Shebuti and Akoglu, Leman},
  booktitle={Proceedings of the 21th ACM SIGKDD international conference on knowledge discovery and data mining},
  pages={985--994},
  year={2015}
}

@inproceedings{hooi2016fraudar,
  title={FRAUDAR: Bounding Graph Fraud in the Face of Camouflage},
  author={Hooi, Bryan and Song, Hyun Ah and Beutel, Alex and Shah, Neil and Shin, Kijung and Faloutsos, Christos},
  booktitle={Proceedings of the 22nd ACM SIGKDD international conference on knowledge discovery and data mining},
  pages={895--904},
  year={2016}
}

@article{kipf2016semi,
  title={Semi-supervised classification with graph convolutional networks},
  author={Kipf, Thomas N and Welling, Max},
  journal={arXiv preprint arXiv:1609.02907},
  year={2016}
}

@inproceedings{velivckovic2018graph,
  title={Graph attention networks},
  author={Veli{\v{c}}kovi{\'c}, Petar and Cucurull, Guillem and Casanova, Arantxa and Romero, Adriana and Lio, Pietro and Bengio, Yoshua and others},
  booktitle={International conference on learning representations},
  volume={6},
  number={2},
  year={2018},
  organization={Ithaca}
}

@article{hamilton2017inductive,
  title={Inductive representation learning on large graphs},
  author={Hamilton, Will and Ying, Zhitao and Leskovec, Jure},
  journal={Advances in neural information processing systems},
  volume={30},
  year={2017}
}

@inproceedings{wang2019semi,
  title={A semi-supervised graph attentive network for financial fraud detection},
  author={Wang, Daixin and Lin, Jianbin and Cui, Peng and Jia, Quanhui and Wang, Zhen and Fang, Yanming and Yu, Quan and Zhou, Jun and Yang, Shuang and Qi, Yuan},
  booktitle={2019 IEEE international conference on data mining (ICDM)},
  pages={598--607},
  year={2019},
  organization={IEEE}
}

@inproceedings{dou2020enhancing,
  title={Enhancing graph neural network-based fraud detectors against camouflaged fraudsters},
  author={Dou, Yingtong and Liu, Zhiwei and Sun, Li and Deng, Yutong and Peng, Hao and Yu, Philip S},
  booktitle={Proceedings of the 29th ACM international conference on information \& knowledge management},
  pages={315--324},
  year={2020}
}

@inproceedings{liu2020alleviating,
  title={Alleviating the inconsistency problem of applying graph neural network to fraud detection},
  author={Liu, Zhiwei and Dou, Yingtong and Yu, Philip S and Deng, Yutong and Peng, Hao},
  booktitle={Proceedings of the 43rd international ACM SIGIR conference on research and development in information retrieval},
  pages={1569--1572},
  year={2020}
}

@inproceedings{liu2021pick,
  title={Pick and choose: a GNN-based imbalanced learning approach for fraud detection},
  author={Liu, Yang and Ao, Xiang and Qin, Zidi and Chi, Jianfeng and Feng, Jinghua and Yang, Hao and He, Qing},
  booktitle={Proceedings of the web conference 2021},
  pages={3168--3177},
  year={2021}
}

@inproceedings{he2024harnessing,
  title={Harnessing explanations: Llm-to-lm interpreter for enhanced text-attributed graph representation learning},
  author={He, Xiaoxin and Bresson, Xavier and Laurent, Thomas and Perold, Adam and LeCun, Yann and Hooi, Bryan},
  booktitle={International conference on learning representations},
  volume={2024},
  pages={5711--5732},
  year={2024}
}

@inproceedings{zhu2024touchup,
  title={Touchup-g: Improving feature representation through graph-centric finetuning},
  author={Zhu, Jing and Song, Xiang and Ioannidis, Vassilis and Koutra, Danai and Faloutsos, Christos},
  booktitle={Proceedings of the 47th International ACM SIGIR Conference on Research and Development in Information Retrieval},
  pages={2662--2666},
  year={2024}
}

@inproceedings{tang2024graphgpt,
  title={Graphgpt: Graph instruction tuning for large language models},
  author={Tang, Jiabin and Yang, Yuhao and Wei, Wei and Shi, Lei and Su, Lixin and Cheng, Suqi and Yin, Dawei and Huang, Chao},
  booktitle={Proceedings of the 47th International ACM SIGIR Conference on Research and Development in Information Retrieval},
  pages={491--500},
  year={2024}
}

@inproceedings{yang2025flag,
  title={Flag: Fraud detection with llm-enhanced graph neural network},
  author={Yang, Chengdong and Liu, Hongrui and Wang, Daixin and Zhang, Zhiqiang and Yang, Cheng and Shi, Chuan},
  booktitle={Proceedings of the 31st ACM SIGKDD Conference on Knowledge Discovery and Data Mining V. 2},
  pages={5150--5160},
  year={2025}
}

@inproceedings{huang2025can,
  title={Can llms find fraudsters? multi-level llm enhanced graph fraud detection},
  author={Huang, Tairan and Wang, Yili and Li, Qiutong and He, Changlong and Gao, Jianliang},
  booktitle={Proceedings of the 33rd ACM International Conference on Multimedia},
  pages={1530--1538},
  year={2025}
}

@article{xiang2025enhancing,
  title={Enhancing attribute-driven fraud detection with risk-aware graph representation},
  author={Xiang, Sheng and Zhang, Guibin and Cheng, Dawei and Zhang, Ying},
  journal={IEEE Transactions on Knowledge and Data Engineering},
  year={2025},
  publisher={IEEE}
}

@inproceedings{li2026dgp,
  title={DGP: A Dual-Granularity Prompting Framework for Fraud Detection with Graph-Enhanced LLMs},
  author={Li, Yuan and Hu, Jun and Hooi, Bryan and He, Bingsheng and Chen, Cheng},
  booktitle={Proceedings of the AAAI Conference on Artificial Intelligence},
  volume={40},
  number={18},
  pages={15171--15179},
  year={2026}
}

@inproceedings{mcauley2013amateurs,
  title={From amateurs to connoisseurs: modeling the evolution of user expertise through online reviews},
  author={McAuley, Julian John and Leskovec, Jure},
  booktitle={Proceedings of the 22nd international conference on World Wide Web},
  pages={897--908},
  year={2013}
}

@inproceedings{weng2019cats,
  title={Cats: cross-platform e-commerce fraud detection},
  author={Weng, Haiqin and Ji, Shouling and Duan, Fuzheng and Li, Zhao and Chen, Jianhai and He, Qinming and Wang, Ting},
  booktitle={2019 ieee 35th international conference on data engineering (icde)},
  pages={1874--1885},
  year={2019},
  organization={IEEE}
}

@inproceedings{tang2022rethinking,
  title={Rethinking graph neural networks for anomaly detection},
  author={Tang, Jianheng and Li, Jiajin and Gao, Ziqi and Li, Jia},
  booktitle={International conference on machine learning},
  pages={21076--21089},
  year={2022},
  organization={PMLR}
}

@inproceedings{li2025diffgraph,
  title={Diffgraph: Heterogeneous graph diffusion model},
  author={Li, Zongwei and Xia, Lianghao and Hua, Hua and Zhang, Shijie and Wang, Shuangyang and Huang, Chao},
  booktitle={Proceedings of the Eighteenth ACM International conference on web search and data mining},
  pages={40--49},
  year={2025}
}

@inproceedings{wang2019fdgars,
  title={Fdgars: Fraudster detection via graph convolutional networks in online app review system},
  author={Wang, Jianyu and Wen, Rui and Wu, Chunming and Huang, Yu and Xiong, Jian},
  booktitle={Companion proceedings of the 2019 World Wide Web conference},
  pages={310--316},
  year={2019}
}

@inproceedings{zhang2019key,
  title={Key player identification in underground forums over attributed heterogeneous information network embedding framework},
  author={Zhang, Yiming and Fan, Yujie and Ye, Yanfang and Zhao, Liang and Shi, Chuan},
  booktitle={Proceedings of the 28th ACM international conference on information and knowledge management},
  pages={549--558},
  year={2019}
}

@inproceedings{liu2018heterogeneous,
  title={Heterogeneous graph neural networks for malicious account detection},
  author={Liu, Ziqi and Chen, Chaochao and Yang, Xinxing and Zhou, Jun and Li, Xiaolong and Song, Le},
  booktitle={Proceedings of the 27th ACM international conference on information and knowledge management},
  pages={2077--2085},
  year={2018}
}

@article{zhuo2024partitioning,
  title={Partitioning message passing for graph fraud detection},
  author={Zhuo, Wei and Liu, Zemin and Hooi, Bryan and He, Bingsheng and Tan, Guang and Fathony, Rizal and Chen, Jia},
  journal={arXiv preprint arXiv:2412.00020},
  year={2024}
}

@inproceedings{yu2024barely,
  title={Barely supervised learning for graph-based fraud detection},
  author={Yu, Hang and Liu, Zhengyang and Luo, Xiangfeng},
  booktitle={Proceedings of the AAAI conference on artificial intelligence},
  volume={38},
  number={15},
  pages={16548--16557},
  year={2024}
}

@article{chang2024survey,
  title={A survey on evaluation of large language models},
  author={Chang, Yupeng and Wang, Xu and Wang, Jindong and Wu, Yuan and Yang, Linyi and Zhu, Kaijie and Chen, Hao and Yi, Xiaoyuan and Wang, Cunxiang and Wang, Yidong and others},
  journal={ACM transactions on intelligent systems and technology},
  volume={15},
  number={3},
  pages={1--45},
  year={2024},
  publisher={ACM New York, NY}
}

@article{thirunavukarasu2023large,
  title={Large language models in medicine},
  author={Thirunavukarasu, Arun James and Ting, Darren Shu Jeng and Elangovan, Kabilan and Gutierrez, Laura and Tan, Ting Fang and Ting, Daniel Shu Wei},
  journal={Nature medicine},
  volume={29},
  number={8},
  pages={1930--1940},
  year={2023},
  publisher={Nature Publishing Group US New York}
}

@inproceedings{wei2024llmrec,
  title={Llmrec: Large language models with graph augmentation for recommendation},
  author={Wei, Wei and Ren, Xubin and Tang, Jiabin and Wang, Qinyong and Su, Lixin and Cheng, Suqi and Wang, Junfeng and Yin, Dawei and Huang, Chao},
  booktitle={Proceedings of the 17th ACM international conference on web search and data mining},
  pages={806--815},
  year={2024}
}

@article{tan2023walklm,
  title={Walklm: A uniform language model fine-tuning framework for attributed graph embedding},
  author={Tan, Yanchao and Zhou, Zihao and Lv, Hang and Liu, Weiming and Yang, Carl},
  journal={Advances in neural information processing systems},
  volume={36},
  pages={13308--13325},
  year={2023}
}

@inproceedings{lu2022bright,
  title={Bright-graph neural networks in real-time fraud detection},
  author={Lu, Mingxuan and Han, Zhichao and Rao, Susie Xi and Zhang, Zitao and Zhao, Yang and Shan, Yinan and Raghunathan, Ramesh and Zhang, Ce and Jiang, Jiawei},
  booktitle={Proceedings of the 31st ACM international conference on information \& knowledge management},
  pages={3342--3351},
  year={2022}
}

@inproceedings{liu2019geniepath,
  title={Geniepath: Graph neural networks with adaptive receptive paths},
  author={Liu, Ziqi and Chen, Chaochao and Li, Longfei and Zhou, Jun and Li, Xiaolong and Song, Le and Qi, Yuan},
  booktitle={Proceedings of the AAAI conference on artificial intelligence},
  volume={33},
  number={01},
  pages={4424--4431},
  year={2019}
}

@inproceedings{shi2022h2,
  title={H2-fdetector: A gnn-based fraud detector with homophilic and heterophilic connections},
  author={Shi, Fengzhao and Cao, Yanan and Shang, Yanmin and Zhou, Yuchen and Zhou, Chuan and Wu, Jia},
  booktitle={Proceedings of the ACM web conference 2022},
  pages={1486--1494},
  year={2022}
}

@inproceedings{wang2020gcn,
  title={Am-gcn: Adaptive multi-channel graph convolutional networks},
  author={Wang, Xiao and Zhu, Meiqi and Bo, Deyu and Cui, Peng and Shi, Chuan and Pei, Jian},
  booktitle={Proceedings of the 26th ACM SIGKDD International conference on knowledge discovery \& data mining},
  pages={1243--1253},
  year={2020}
}

@article{zhu2020deep,
  title={Deep graph contrastive representation learning},
  author={Zhu, Yanqiao and Xu, Yichen and Yu, Feng and Liu, Qiang and Wu, Shu and Wang, Liang},
  journal={arXiv preprint arXiv:2006.04131},
  year={2020}
}

@article{velivckovic2018deep,
  title={Deep graph infomax},
  author={Veli{\v{c}}kovi{\'c}, Petar and Fedus, William and Hamilton, William L and Li{\`o}, Pietro and Bengio, Yoshua and Hjelm, R Devon},
  journal={arXiv preprint arXiv:1809.10341},
  year={2018}
}

@inproceedings{ye2024language,
  title={Language is all a graph needs},
  author={Ye, Ruosong and Zhang, Caiqi and Wang, Runhui and Xu, Shuyuan and Zhang, Yongfeng},
  booktitle={Findings of the association for computational linguistics: EACL 2024},
  pages={1955--1973},
  year={2024}
}

@article{minaee2024large,
  title={Large language models: A survey},
  author={Minaee, Shervin and Mikolov, Tomas and Nikzad, Narjes and Chenaghlu, Meysam and Socher, Richard and Amatriain, Xavier and Gao, Jianfeng},
  journal={arXiv preprint arXiv:2402.06196},
  year={2024}
}

@article{zhao2022learning,
  title={Learning on large-scale text-attributed graphs via variational inference},
  author={Zhao, Jianan and Qu, Meng and Li, Chaozhuo and Yan, Hao and Liu, Qian and Li, Rui and Xie, Xing and Tang, Jian},
  journal={arXiv preprint arXiv:2210.14709},
  year={2022}
}

@article{achiam2023gpt,
  title={Gpt-4 technical report},
  author={Achiam, Josh and Adler, Steven and Agarwal, Sandhini and Ahmad, Lama and Akkaya, Ilge and Aleman, Florencia Leoni and Almeida, Diogo and Altenschmidt, Janko and Altman, Sam and Anadkat, Shyamal and others},
  journal={arXiv preprint arXiv:2303.08774},
  year={2023}
}

@article{yang2025qwen3,
  title={Qwen3 technical report},
  author={Yang, An and Li, Anfeng and Yang, Baosong and Zhang, Beichen and Hui, Binyuan and Zheng, Bo and Yu, Bowen and Gao, Chang and Huang, Chengen and Lv, Chenxu and others},
  journal={arXiv preprint arXiv:2505.09388},
  year={2025}
}

@article{grattafiori2024llama,
  title={The llama 3 herd of models},
  author={Grattafiori, Aaron and Dubey, Abhimanyu and Jauhri, Abhinav and Pandey, Abhinav and Kadian, Abhishek and Al-Dahle, Ahmad and Letman, Aiesha and Mathur, Akhil and Schelten, Alan and Vaughan, Alex and others},
  journal={arXiv preprint arXiv:2407.21783},
  year={2024}
}

\end{document}